%% file: flickr30k_entities.tex
\documentclass[twocolumn]{svjour3} 
\smartqed 

\usepackage[table,x11names]{xcolor}
\usepackage{soul}

\usepackage[round]{natbib}
\setcitestyle{aysep={}}
\usepackage{floatrow}
\DeclareFloatFont{tiny}{\tiny}
\usepackage{times}
\usepackage{epsfig}
\usepackage{graphicx}
\usepackage{amsmath}
\usepackage{amssymb}
\usepackage{multirow}
\usepackage{paralist}
\usepackage{verbatim}
\usepackage{stackengine}
\usepackage{graphicx}
\usepackage[export]{adjustbox}

\usepackage[pagebackref=false,breaklinks=true,letterpaper=true,colorlinks,bookmarks=false,hidelinks]{hyperref}

\usepackage{algorithm2e}

\usepackage{svg}
\setsvg{svgpath = Figures/}

\begin{document}

\title{Flickr30k Entities: Collecting Region-to-Phrase Correspondences for Richer Image-to-Sentence Models}


\author{Bryan A. Plummer         
\and
Liwei Wang
\and 
Chris M. Cervantes
\and 
Juan C. Caicedo
\and 
Julia Hockenmaier
\and 
Svetlana Lazebnik
}


\institute{
B. A. Plummer \at
University of Illinois at Urbana Champaign, Urbana, IL, USA\\
\email{bplumme2@illinois.edu}          
\and
L. Wang \at
University of Illinois at Urbana Champaign, Urbana, IL, USA
\and
C. M. Cervantes \at
University of Illinois at Urbana Champaign, Urbana, IL, USA
\and
J. C. Caicedo \at
Broad Institute of MIT and Harvard, Boston, MA, USA
 \and
J. Hockenmaier \at
University of Illinois at Urbana Champaign, Urbana, IL, USA
\and
S. Lazebnik \at
University of Illinois at Urbana Champaign, Urbana, IL, USA
}

\date{Received: date / Accepted: date}

\maketitle

\begin{abstract}
The Flickr30k dataset has become a standard benchmark for sentence-based image description. This paper presents Flickr30k Entities, which augments the 158k captions from Flickr30k with 244k coreference chains, linking mentions of the same entities across different captions for the same image, and associating them with 276k manually annotated bounding boxes. Such annotations are essential for continued progress in automatic image description and grounded language understanding. They enable us to define a new benchmark for localization of textual entity mentions in an image.  We present a strong baseline for this task that combines an image-text embedding, detectors for common objects, a color classifier, and a bias towards selecting larger objects. While our baseline rivals in accuracy more complex state-of-the-art models, we show that its gains cannot be easily parlayed into improvements on such tasks as image-sentence retrieval, thus underlining the limitations of current methods and the need for further research.
\keywords{Computer Vision \and Language \and Region Phrase Correspondence \and Datasets \and Crowdsourcing}
\end{abstract}

\begin{figure*}[!tb]
\centering
\begin{tabular}[t]{ccc}
\includegraphics[height=1.75in]{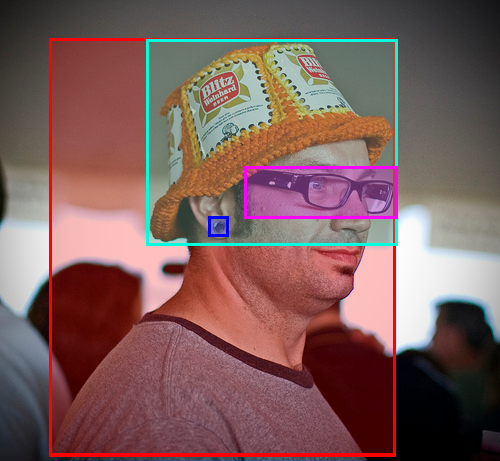} \vspace{-3mm}& 
\includegraphics[height=1.75in]{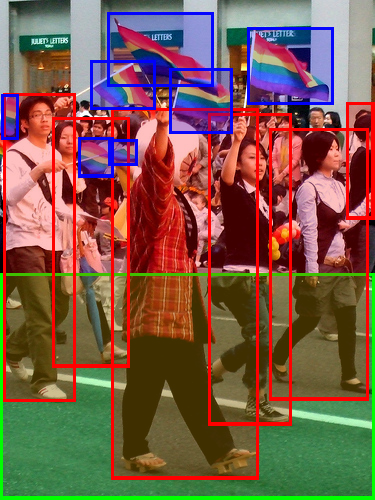} &
\includegraphics[height=1.75in]{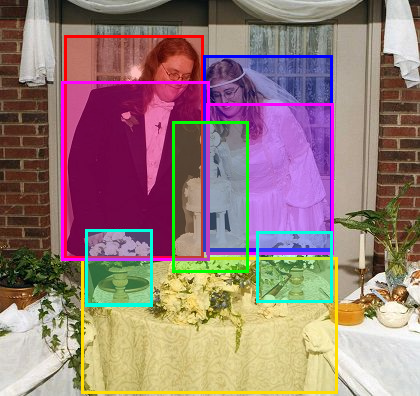} \\ 
\includegraphics[valign=T,width=.31\columnwidth]{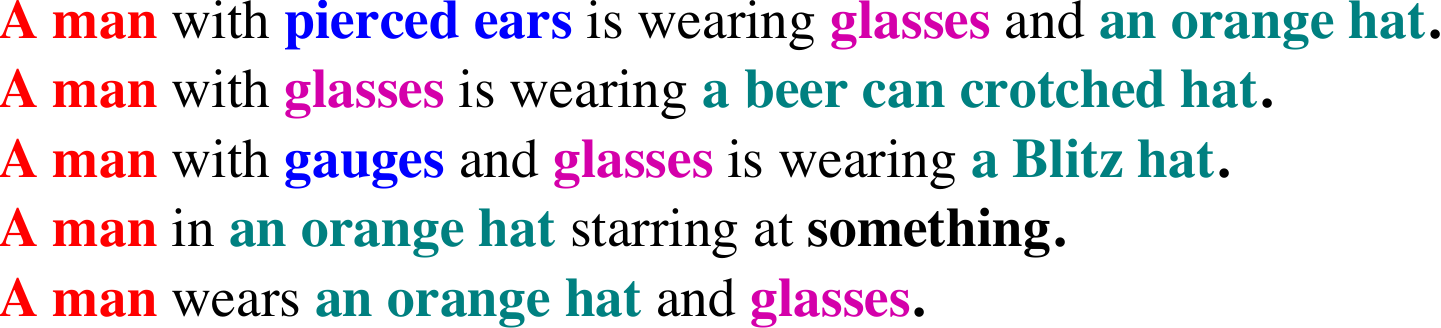}&
\includegraphics[valign=T,width=.31\columnwidth]{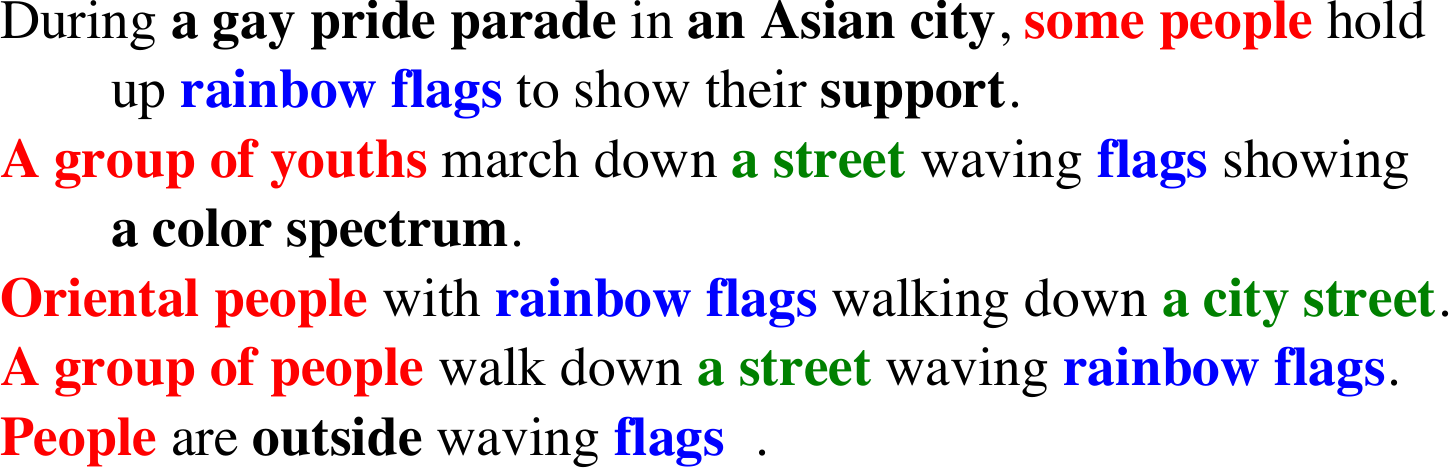}&
\includegraphics[valign=T,width=.27\columnwidth]{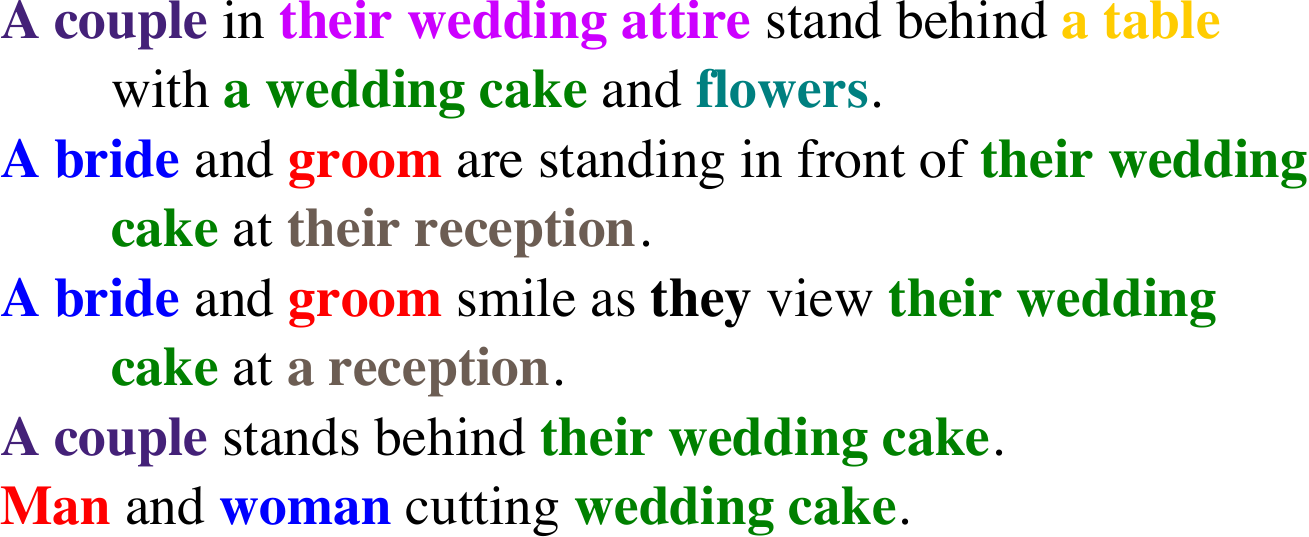}\\
\end{tabular}
\caption{Example annotations from our dataset. In each group of captions describing the same image, coreferent mentions ({\em coreference chains}) and their corresponding bounding boxes are marked with the same color. On the left, each chain points to a single entity (bounding box). Scenes and events like ``outside" or ``parade" have no box. In the middle example, the people (red) and flags (blue) chains point to multiple boxes each. On the right,  blue phrases refer to the bride, and red phrases refer to the groom. The dark purple phrases (``a couple'') refer to both of these entities, and their corresponding bounding boxes are identical to the red and blue ones.}
    \label{fig:Examples}
\end{figure*}

\section{Introduction}

From robotics to human-computer interaction, there are numerous real-world tasks that would benefit from practical, large-scale systems that can identify objects in scenes based on language and understand language based on visual context. There has been a recent surge of work in this area, and in particular, on the task of sentence-based image description~\citep{chen2014learning,donahue2014long,fang2014captions,Farhadi10,Hodosh13,karpathy2014deep,kiros2014unifying,klein2014fisher,Kulkarni11,lebret2015phrase,mao2014deep,Ordonez11,vinyals2014show,Yao2010} and visual question answering~\citep{VQA,gao2015mQA,krishnavisualgenome,malinowski2014nips,ren_cocoqa,VisualMadlibs}. Unfortunately, due to a lack of datasets that provide not only paired sentences and images, but detailed {\em grounding} of specific phrases in image regions, most of these methods attempt to directly learn mappings from whole images to whole sentences. Not surprisingly, such models have a tendency to reproduce generic captions from the training data, and to perform poorly on {\em compositionally novel} images whose objects may have been seen individually at training time, but not in that combination~\citep{Devlin15}. Some recent works do try to find correspondences between image regions and parts of sentences~\citep{fang2014captions,karpathy2014deepNIPS,karpathy2014deep,xu2015show}, but they treat such correspondences as latent and do not evaluate their quality directly. But this paper argues, and our own preliminary results indicate, that grounding of language to image regions is a problem that is hard and fundamental enough to require more extensive ground-truth annotations and standalone benchmarks. 


The main contribution of this paper is providing a large-scale comprehensive dataset of region-to-phrase correspondences for image description. We build on the Flickr30k dataset~\citep{young2014image}, a popular benchmark for caption generation and retrieval that has been used, among others, by~\cite{chen2014learning,donahue2014long,fang2014captions,gong2014improving,karpathy2014deepNIPS,karpathy2014deep,kiros2014unifying,klein2014fisher,lebret2015phrase,mao2014deep,vinyals2014show,xu2015show}.  Flickr30k contains 31,783 images focusing mainly on people and animals, and 158,915 English captions (five per image). Our new dataset, Flickr30k Entities, augments Flickr30k by identifying which mentions among the captions of the same image refer to the same set of entities, resulting in 244,035 {\em coreference chains}, and which image regions depict the mentioned entities, resulting in 275,775 bounding boxes. Figure \ref{fig:Examples} illustrates the structure of our annotations on three sample images. Section \ref{sec:data} describes our crowdsourcing protocol, which consists of two major stages -- coreference resolution and bounding box drawing -- and each stage in turn is split up into smaller atomic tasks to ensure both efficiency and quality.

Together with our annotations, we propose a new benchmark task of {\em phrase localization}, which we view as a fundamental building block and prerequisite for more advanced image-language understanding tasks. Given an image and a caption that accurately describes it, the goal of phrase localization is to predict a bounding box for a specific entity mention from that sentence. This task is akin to object detection and can in principle be evaluated in an analogous way, but it has its own unique challenges. Traditional object detection assumes a predefined list of semantically distinct classes with many training examples for each. By contrast, in phrase localization, the number of possible phrases is very large, and many of them have just a single example or are completely unseen at training time. Also, different phrases may be very semantically similar (e.g., \emph{infant} and \emph{baby}), which makes it difficult to train separate models for each. And of course, to deal with the full complexity of this task, we need to take into account the broader context of the whole image and sentence, for example, when disambiguating between multiple entities of the same type. In Section \ref{sec:results}, we propose a strong baseline for this task based on a combination of image-text embeddings, pre-trained detectors, and size and color cues. While this baseline outperforms more complex recent methods (e.g.,~\cite{rohrbach2015}), it is not yet strong enough to discriminate between multiple competing interpretations that roughly fit an image, which is necessary to achieve improvements over state-of-the-art global methods for image description.


A preliminary version of this work has appeared in~\cite{flickrentities}. The present journal paper includes a more detailed description and analysis of our crowdsourcing protocol, as well as brand new, much stronger baseline results. By using better region features (Fast RCNN~\citep{girshickICCV15fastrcnn} instead of ImageNet-trained VGG~\citep{simonyan2014very}) in combination with size and color cues, we are able to improve the Recall@1 for phrase localization from approximately 25\% to 50\% (Section \ref{sec:localization}).

Our dataset is available for download at \url{http://web.engr.illinois.edu/~bplumme2/Flickr30kEntities/}

\begin{table*}
\caption{Comparison of dataset statistics. For our dataset, we define Object Categories as the set of unique phrases after filtering out non-nouns in our annotated phrases (note that Scene Graph and Visual Genome also have very large numbers in this column because they correspond essentially to the total numbers of unique phrases). For Expressions Per Image, we list for our dataset the average number of entity mentions in all five sentences. }
\label{table:dataCompare}
\centering
\begin{tabular}{|l|l|r|r|r|r|r|r|r|r|}
\hline
& \multirow{2}{*}{Dataset} & \multirow{2}{*}{Images} &Objects Per & Object & Objects Per & Sentences & Expressions\\
&  & & Image & Categories &Category &  Per Image & Per Image\\
\hline
Image-Sentence & {\bf Flickr30k Entities} & 31,783 & 8.7 & 44,518 & 6.2 & 5 & 16.6 \\
Datasets & MSCOCO~\citep{lin2014microsoft} & 328,000 & 7.7 & 91 & 27,473 & 5 & -- \\
 \hline
 & ReferIt~\citep{kazemzadeh-EtAl:2014:EMNLP2014} & 19,894 & 4.9 & 238 & 406.1 & -- & 6.6\\
Image-Phrase & Google Refexp~\citep{mao2015generation} & 26,711 & 2.1 & 80 & 685.3 & -- & 3.9\\
Datasets & Scene Graph~\citep{Johnson2015CVPR} & 5,000 & 18.8 & 6,745 & 13.9 & -- & 33.0\\
& Visual Genome\textsuperscript{*}~\citep{krishnavisualgenome} & 108,077 & $\sim$56 & 110,689 & $\sim$54 & -- & $\sim$40 \\
\hline
\end{tabular}
\vspace{-3mm}
\flushleft\hspace{1mm}\begingroup
    \fontsize{6pt}{6pt}\selectfont
 *obtained via personal communication with the authors
\endgroup
\vspace{-3mm}
\end{table*}

\section{Related Work}
\label{relatedWork}

\subsection{Datasets with Region-Level Descriptions}
\label{sec:datasets}

We begin our discussion with datasets that pair images with global text descriptions and also include some kind of region-level annotations. An early example of this is the UIUC Sentences dataset~\citep{rashtchian2010collecting}, which consists of 1,000 images from PASCAL VOC 2008~\citep{pascal-voc-2008} and five sentences per image. It inherits from PASCAL object annotations for 20 categories, but lacks explicit links between its captions and the object annotations. The most recent and large-scale dataset of this kind is Microsoft Common Objects in Context (MSCOCO)~\citep{lin2014microsoft}, containing over 300k images with five sentences per image and over 2.5m labeled object instances from 91 pre-defined categories. However, just as in UIUC Sentences, the MSCOCO region-level annotations are not linked to the captions in any way.

Rather than pairing images with a caption that summarizes the entire image, some datasets pair specific objects in an image with short descriptions. The ReferIt dataset~\citep{kazemzadeh-EtAl:2014:EMNLP2014} focuses on {\em referring expressions} that are necessary to uniquely identify an object instance in an image. It augments the IAPR-TC dataset~\citep{grubinger2006iapr} of 20k photographs  with 130k isolated entity descriptions for 97k objects from 238 categories. The Google Refexp dataset~\citep{mao2015generation} is built on top of MSCOCO and contains a little under 27k images with 105k descriptions, and it uses a methodology that produces longer descriptions than ReferIt.
Visual MadLibs~\citep{VisualMadlibs} is a subset of 10,738 MSCOCO images with several types of focused fill-in-the-blank descriptions (360k in total), some referring to attributes and actions of specific people and object instances, and some referring to the image as a whole.

\citet{Johnson2015CVPR} is another notable work concerned with grounding of semantic scene descriptions to image regions. Instead of natural language, it proposes a formal {\em scene graph} representation that encapsulates all entities, attributes and relations in an image, together with a dataset of scene graphs and their groundings for 5k images. 
The more recent Visual Genome dataset~\citep{krishnavisualgenome} follows the same methodology, but contains 108k images rather than 5k and a denser set of annotations. Each image in Visual Genome has an average of 21 objects, 18 attributes, and 18 pairwise relations. Due to the nature of the Visual Genome crowdsourcing protocol, its object annotations have a greater amount of redundancy than our dataset. For example, the phrases {\em a boy wearing jeans} and {\em this is a little boy} may be totally separate and come with separate bounding boxes despite referring to the same person in the image. In addition, for phrases referring to multiple objects like \emph{three people}, Visual Genome would only have one box drawn around all three people, while we asked for individual boxes for each person, linking all three boxes to the phrase. While the Visual Genome is the largest source of unstructured localized textual expressions to date, our dataset is better suited for understanding the different ways people refer to the same visual entities within an image, and which entities are salient for the purpose of natural language description.

Finally, there exist a few specialized datasets with extensive annotations, but more limited domains of applicability than Flickr30k Entities or Visual Genome. \citet{kong2014you} have taken the 1,449 RGB-D images of static indoor scenes from the NYUv2 dataset~\citep{Silberman:ECCV12} and obtained detailed multi-sentence descriptions focusing mainly on spatial relationships between objects.  Similar to Flickr30k Entities, this dataset contains links between different mentions of the same object, and between words in the description and the respective location in the image. \cite{abstractscene} have introduced the Abstract Scene dataset, which contains 10,020 synthetic images created using clip art objects from 58 categories, together with captions and ground-truth information of how objects relate to the captions. 

Table \ref{table:dataCompare} compares the statistics of Flickr30k Entities with key related datasets.

\subsection{Grounded Language Understanding}

As mentioned in the Introduction, the most common image-language understanding task in the literature is automatic image captioning~\citep{chen2014learning,donahue2014long,fang2014captions,Farhadi10,Hodosh13,karpathy2014deep,kiros2014unifying,klein2014fisher,klein2015rnn,Kulkarni11,lebret2015phrase,ma2015,mao2014deep,Ordonez11,vinyals2014show,Yao2010}. Of most importance to us are the methods attempting to associate local regions in an image with words or phrases in the captions, as they would likely benefit the most from our annotations.

Many works leveraging region-phrase correspondences rely on weakly supervised learning due to a lack of ground-truth correspondences at training time.  \citet{fang2014captions} use multiple instance learning to train detectors for words that commonly occur in captions, and then feed the outputs of these detectors into a language model to generate novel captions. \citet{xu2015show} incorporate a soft form of attention into their recurrent model, which is trained to fixate on a sequence of latent image regions while generating words. \citet{karpathy2014deepNIPS,karpathy2014deep} propose an image-sentence ranking approach in which the score between an image and sentence is defined as the average over correspondence scores between each sentence fragment and the best corresponding image region; at training time, the correspondences are treated as latent and incorporated into a structured objective. \citet{ma2015} learn multiple networks capturing word, phrase, and sentence-level interactions with an image and combine the scores of these networks to obtain a whole image-sentence score.  Since there is no explicit mapping between phrases and the image, all three networks use the whole image representation as input. 

In this paper, we mostly step back from the task of whole-image description, and as a prerequisite for it, consider the task of grounding or localizing textual mentions of entities in an image. Until recently, it was rare to see direct evaluation on this task due to a lack of ground-truth annotations (with the notable exception of \citet{kong2014you} and their dataset of RGB-D room descriptions). \cite{rohrbach2015} were among the first to use Flickr30K Entities for phrase localization by training an LSTM model to attend to the right image region in order to reproduce a given phrase. Their work shows that fully supervised training of this model with ground-truth region-phrase correspondences results in much better performance than weakly supervised training, thus confirming the usefulness of our annotations. Since then, a number of other works have adopted Flickr30K Entities as well. \citet{hu2016natural} leverage spatial information and global context to model where objects are likely to occur. \citet{deepspite2015} learn a nonlinear region-phrase embedding that can localize phrases more accurately than our linear CCA embedding of Section \ref{sec:results}. \citet{wang2016matching} formulate a linear program to localize all the phrases from a caption jointly, taking their semantic relationships into account.  \citet{zhang2016EB} perform phrase localization with a tag prediction network and a top-down attention model.

As a more open-ended alternative to phrase localization, \citet{densecap2015} introduce dense image captioning, or the task of predicting image regions and generating freeform descriptions for them, together with a neural network model for this task trained on the Visual Genome dataset~\citep{krishnavisualgenome}.

\input{data}

\input{phraseLocalization}

\input{retrieval}

\section{Conclusion}

This paper has presented Flickr30k Entities, a large-scale image description dataset that provides comprehensive ground-truth correspondence between regions in images and phrases in captions. Our annotations can be used to benchmark tasks like phrase localization, for which up to now large-scale ground-truth information has been lacking. While methods for global image description have been improving rapidly, our experiments suggest that even the current state-of-the-art models still have a limited ability to ground specific textual mentions in local image regions. Methods that make use region-phrase correspondences, e.g., the sentence generation models of \citet{karpathy2014deepNIPS,karpathy2014deep,fang2014captions}, should be able to make use of datasets like ours to continue to make progress on the problem. 

Because our dataset is densely annotated with multiple boxes per image linked to their textual mentions in a larger sentence context, it will also be a rich resource for learning models of multi-object spatial layout~\citep{Farhadi10,fidler2013,Kulkarni11}. Other potential applications include training models for automatic cross-caption coreference~\citep{hodosh2010cross}, distinguishing visual from non-visual text~\citep{Dodge2012}, and estimation of visual saliency for natural language description tasks.  Region-phrase correspondences can also be useful for training better models for visual question answering~\citep{tommasi2016,fukui16emnlp}.

\begin{acknowledgements} This material is based upon work supported by the National Science Foundation under Grants No. 1053856, 1205627, 1405883, 1228082, 1302438, 1563727, as well as support from Xerox UAC and the Sloan Foundation. 
We thank the NVIDIA Corporation for the generous donation of the GPUs used for our experiments.
\end{acknowledgements}

\bibliographystyle{spbasic}
\bibliography{flickr30k_entities.bib}

\end{document}

%% file: data.tex
\section{Annotation Process} \label{sec:data}

\begin{figure*}
\centering
\begin{tabular}[t]{c|c} 
\multicolumn{1}{l|}{\bf(a) Binary Corefernce Link Interface}&
\multicolumn{1}{|l}{\bf(b) Coreference Chain Verification Interface}\\
\raisebox{.1\height}{\includegraphics[width=\textwidth/2,trim=20mm 0 15mm 5mm]{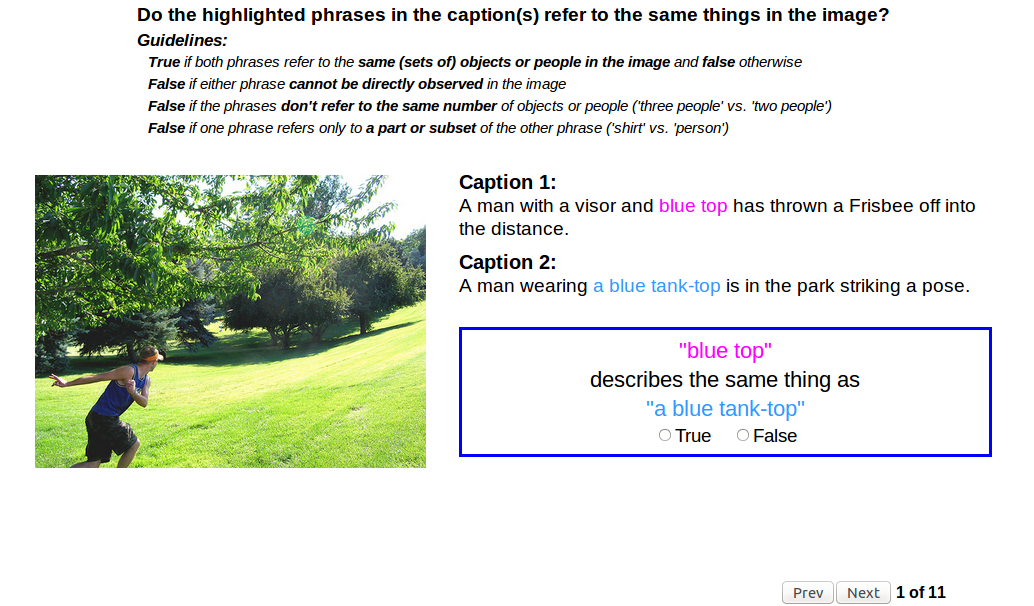}}
 & 
\includegraphics[width=\textwidth/2]{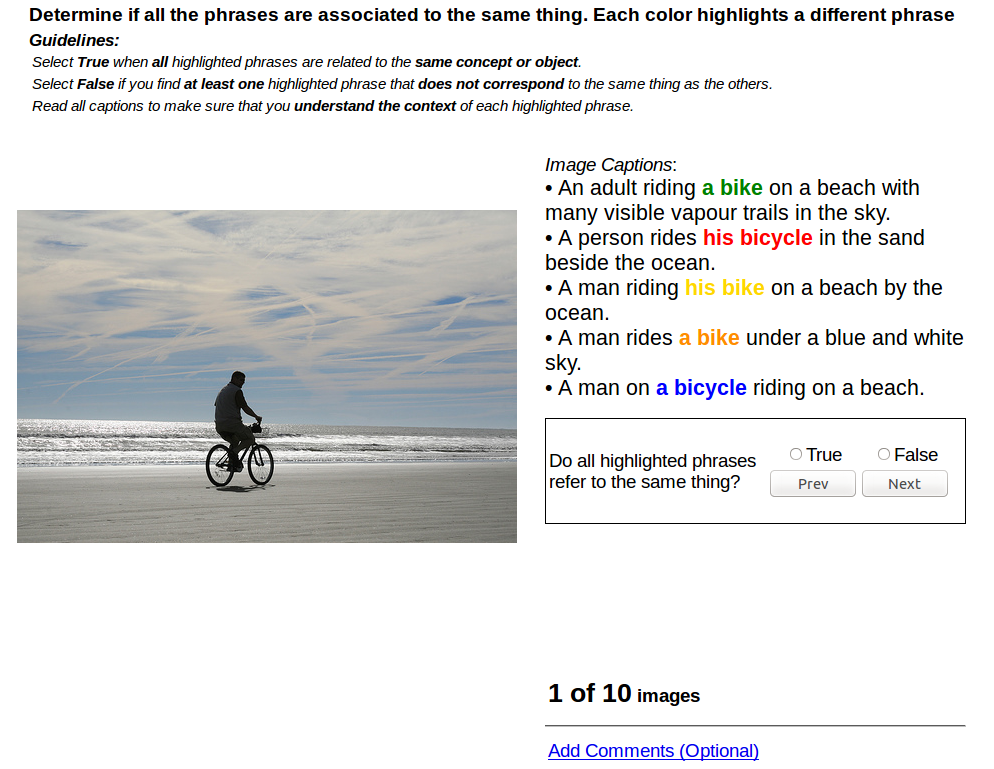} \\
\hline & \\
\multicolumn{1}{l|}{\bf(c) Box Requirement Interface}&
\multicolumn{1}{|l}{\bf(d) Box Drawing Interface}\\
\includegraphics[width=\textwidth/2,trim=0 0 0 -5mm]{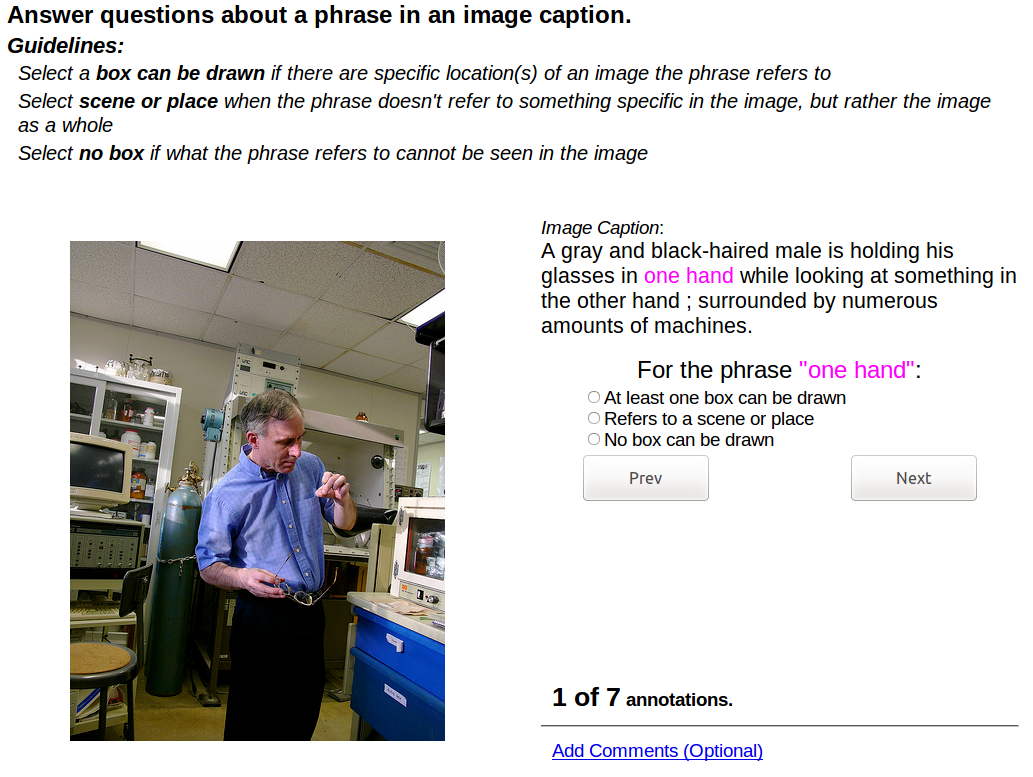} &
\raisebox{.15\height}{\includegraphics[width=\textwidth/2]{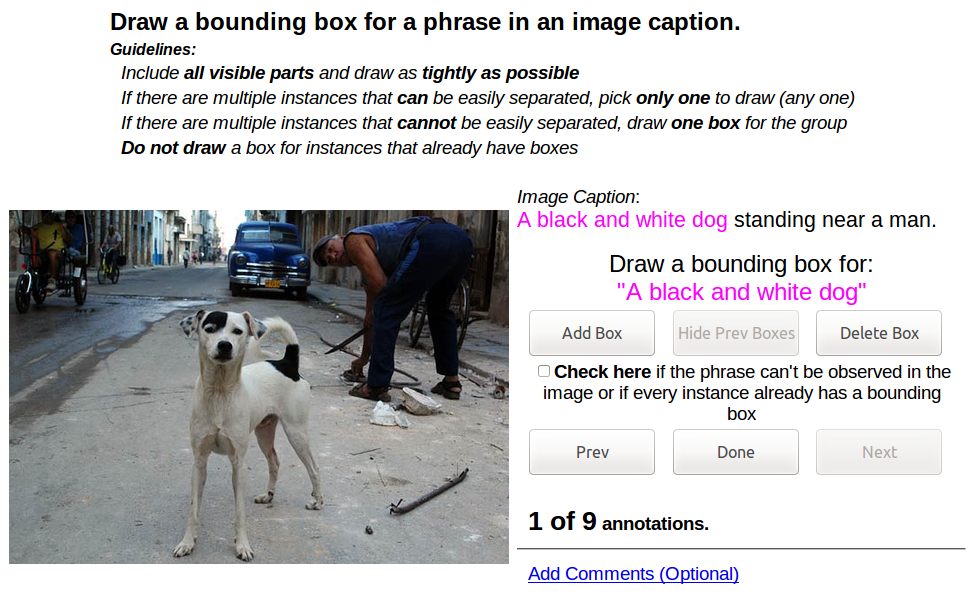}} \\
\hline & \\
\multicolumn{1}{l|}{\bf(e) Box Quality Interface}&
\multicolumn{1}{|l}{\bf(f) Box Coverage Interface}\\
\includegraphics[width=\textwidth/2,trim=0 0 0 -5mm]{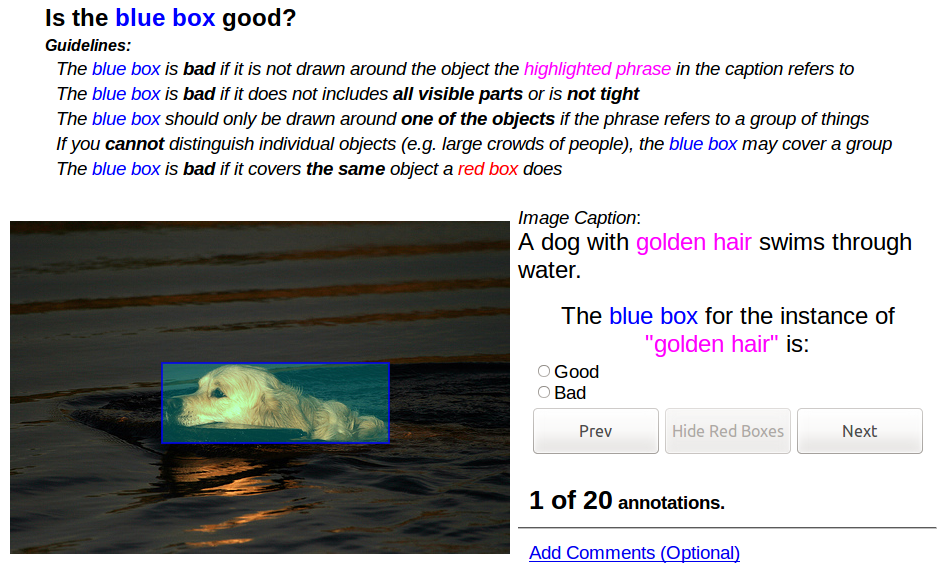} &
\raisebox{.05\height}{\includegraphics[width=\textwidth/2]{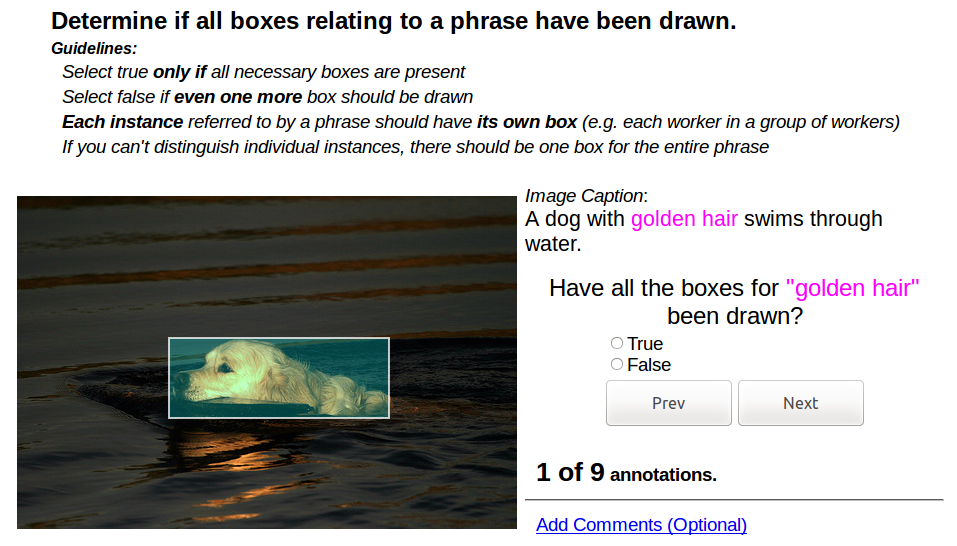}} \\
\end{tabular}
\caption{Examples of the interfaces used in our annotation pipeline described in Section~\ref{sec:data}.}
\label{fig:interfaces}
\end{figure*}

In this section, we describe the crowdsourcing protocol we adopted for collecting Flickr30k Entities. Our annotations, illustrated in Figure \ref{fig:Examples}, consist of cross-caption coreference chains linking mentions of the same entities together with bounding boxes localizing those entities in the image. These annotations are highly structured and vary in complexity from image to image, since images vary in the numbers of clearly distinguishable entities they contain, and sentences vary in the extent of their detail. Further, there are ambiguities involved in identifying whether two mentions refer to the same entity or set of entities, how many boxes (if any) these entities require, and whether these boxes are of sufficiently high quality. Due to this intrinsic subtlety of our task, compounded by the unreliability of crowdsourced judgments, we developed a pipeline of simpler atomic tasks, screenshots of which are shown in Figure~\ref{fig:interfaces}. These tasks can be grouped into two main stages: {\bf coreference resolution}, or forming coreference chains that refer to the same entities (Section \ref{sec:coref}), and {\bf bounding box annotation} for the resulting chains (Section \ref{sec:box}). This workflow provides two advantages: first, identifying coreferent mentions helps reduce redundancy and save box-drawing effort; and second, coreference annotation is intrinsically valuable, e.g., for training cross-caption coreference models~\citep{hodosh2010cross}. Section \ref{sec:quality} will discuss issues connected to data quality, and Section \ref{sec:statistics} will give a brief analysis of dataset statistics.

\subsection{Coreference Resolution}  \label{sec:coref}

We rely on the chunking information given in the Flickr30k captions \citep{young2014image} to identify potential entity mentions. With the exception of personal pronouns (\textit{he, she, they}) and a small list of frequent non-visual terms (\textit{background}, \textit{air}), we assume that any noun-phrase (NP) chunk is a potential entity mention. NP chunks are short (avg. 2.35 words), non-recursive phrases (e.g., the complex NP \textit{[[a man] in [an orange hat]]} is split into two chunks). Mentions may refer to single entities (\textit{a dog}); regions of ``stuff" (\textit{grass}); multiple distinct entities (\textit{two men}, \textit{flags}, \textit{football players}); groups of entities that may not be easily identified as individuals ({\em a crowd}, \textit{a pile of oranges}); or even the entire scene (\textit{the park}). Finally, some NP chunks may not refer to any physical entities (\textit{wedding reception}, \textit{a trick}, \textit{fun}).  

Once we have our candidate mentions from the sentences corresponding to the same image, we need to identify which ones refer to the same set of entities. Since each caption is a single, relatively short sentence, pronouns (\textit{he}, \textit{she}, \textit{they}) are relatively rare in this dataset. Therefore, unlike in standard coreference resolution in running text~\citep{soon2001machine}, which can be beneficial for identifying all mentions of people in movie scripts \citep{RamanathanJLF14}, we ignore anaphoric references between pronouns and their antecedents and focus on cross-caption coreference resolution~\citep{hodosh2010cross}.  Like standard coreference resolution, our task partitions the set of mentions $M$ in a document (here, the five captions of one image), into subsets of equivalent mentions such that all mentions in the same subset $c \in C$ refer to the same set of entities. In keeping with standard terminology, we refer to each such set or cluster of mentions $c \subset M$ as a coreference chain. 

\subsubsection{Binary Coreference Link Annotation}
Since the task of constructing an entire coreference chain from scratch is cognitively complex and error-prone, we broke it down into simpler tasks collecting binary coreference links between pairs of mentions. A coreference link between mentions $m$ and $m^{\prime}$ indicates that $m$ and $m^{\prime}$ refer to the same set of entities. In the manual annotation process, workers are shown an image and the two captions from which $m$ and $m^{\prime}$ originate. The workers are asked whether these mentions refer to the same entity.  See Figure~\ref{fig:interfaces}(a) for a screenshot of the interface for this task. If a worker indicates that the mentions are coreferent, we add a link between $m$ and $m^{\prime}$. Given a set of mentions $M$ for an images, manual annotation of all $O(|M|^{2})$ pairwise links is very costly. But since $M$ typically contains multiple mentions that refer to the same set of entities, the number of coreference chains is bounded by, and typically much smaller than, $|M|$. This allows us to reduce the number of links that need to be annotated to $O(|M||C|)$ by leveraging the transitivity of the coreference relation~\citep{mccarthy1995using}. Given a set of identified coreference chains $C$ and a new mention $m$ that has not been annotated for coreference yet, we only have to ask for links between $m$ and one mention from each element of $C$. If $m$ is not coreferent with any of these mentions, it refers to a new entity whose coreference chain is initialized and added to $C$.

In the worst case, each entity has only one mention requiring annotation of all $|M|^{2}$ possible links. But in practice, most images have more mentions than coreference chains (in our final dataset, each image has an average of 16.6 mentions and 7.8 coreference chains). We further reduce the number of required annotations with two simplifying assumptions.  First, we assume that mentions from the same captions cannot be coreferent, as it would be unlikely for a caption to contain two non-pronominal mentions to the same set of entities. Second, we categorize each mention into eight coarse-grained types using manually constructed dictionaries (people, body parts, animals, clothing/color,\footnote{In Flickr30k, NP chunks that only consist of a color term are often used to refer to clothing, e.g.  \textit{man in blue}.} instruments, vehicles, scene,  and other), and assume mentions belonging to different categories cannot be coreferent. 

\begin{figure}
\centering
\includegraphics[width=0.7\textwidth]{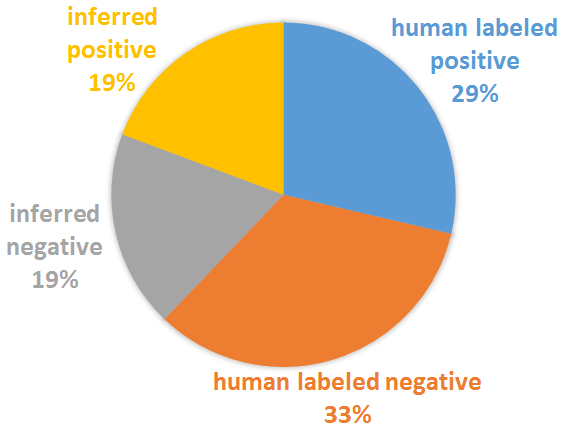}
\caption{Distribution of the source of binary coreference link annotations on the entire dataset using the Greedy Plus strategy.}
\label{fig:binaryLinkSource}
\end{figure}

To ensure that our greedy strategy leveraging the transitivity relations would not have a significant impact on data quality, we conducted a small-scale experiment using 200 images.  First, we asked workers to annotate each of the $O(|M|^{2})$ pairwise links several times to obtain a set of gold (ground-truth) coreference chains.  Then we collected the links again using both the exhaustive and greedy strategies and compared them to the gold links. In addition, after collecting the links, we looked for any violations of transitivity between phrases and asked additional workers to annotate the links involved until we got a consensus. We call the resulting strategies ``exhaustive plus'' and ``greedy plus.'' As seen in the Table~\ref{table:binaryLinkExperiment}, the greedy and exhaustive strategies perform quite similarly, ``greedy plus'' actually performs better than exhaustive while requiring more than 30\% fewer links, and ``exhaustive plus'' achieves the highest accuracy on this task but at prohibitive cost. Based on these considerations, we decided to use ``greedy plus'' for the entire dataset, and Figure~\ref{fig:binaryLinkSource} shows the source of the links we obtained using this strategy. 

\begin{table*}
\caption{Comparison of different annotation strategies for collecting binary coreference links on 200 images.  We report the false positive/negative rates for the individual binary link judgments, as well as how many of the coreference chains created by the different strategies matched the gold coreference chains.}
\label{table:binaryLinkExperiment}
\begin{tabular}{|l|r|r|r|r|}
\hline
 & Exhaustive & Greedy & Exhaustive Plus & Greedy Plus \\
\hline
\hline
Matched Gold Links & 95.03\% & 94.46\% & 98.33\% & 96.40\% \\
Matched Gold Coref. Chain & 81.84\% & 81.55\% & 90.67\% & 86.07\% \\
False Positive Rate & 4.92\% & 5.74\%  & 1.04\% & 3.14\% \\
False Negative Rate & 4.70\% & 4.99\% & 2.28\% & 4.05\% \\
\hline
Links To Annotate & 100.00\% & 57.00\% & 119.74\% & 66.94\%\\
\hline
\end{tabular}
\end{table*}

\subsubsection{Coreference Chain Verification}
To handle errors introduced by the coreference link annotation, we verify the accuracy of all chains that contain more than a single mention. In this task, workers are shown the mentions that belong to the same coreference chain and asked whether all the mentions refer to the same set of entities. If the worker answers True, the chain is kept as-is. If a worker answers False, that chain is broken into subsets of mentions that share the same head noun (the last word in a chunk). An example of the interface for this task is shown in Figure~\ref{fig:interfaces}(b). There were 123,758 coreference chains with more than a single mention to verify in this stage.  Of them, 111,628 (90.2\%) were marked as good by workers, with the remaining 12,130 (9.8\%) marked as bad and broken up before moving on the next step of the annotation pipleline.

It is important to note that our coreference chain verification is not designed to spot false negatives, or missing coreference links. Although false negatives lead to fragmented entities and redundant boxes (and consequently higher time and cost for box drawing), we can recover from many of these errors in a later stage by merging bounding boxes that have significant overlap (Section \ref{sec:mergingprocess}). On the other hand, false positives (spurious coreference links) are more harmful, since they are likely to result in mentions being associated with incorrect entities or image regions.

\subsection{Bounding Box Annotations} 
\label{sec:box}

The workflow to collect bounding box annotations is broken down similarly to~\citet{HJL_AAAI12}, and consists of four separate AMT tasks, discussed below: \begin{inparaenum}[(1)] \item Box Requirement, \item Box Drawing, \item Box Quality, and \item Box Coverage\end{inparaenum}. In each task, workers are shown an image and a caption in which a representative mention for one coreference chain is highlighted. We use the longest mention in each chain, since we assume that it is the most specific.

\subsubsection{Box Requirement}
First, we determine if the entities a representative mention refers to require boxes to be drawn.  A mention does not require boxes if it refers to the entire scene (\textit{in [the park]}), to physical entities that are not in the image (\textit{pose for [the camera]}), or to an action or abstract entity (\textit{perform [a trick]}). As shown in the example interface in Figure~\ref{fig:interfaces}(c), given an image and a caption with a highlighted mention, we ask workers whether
\begin{inparaenum}[(1)]
	\item{at least one box can be drawn}
    \item{the mention refers to a scene or place or}
    \item{no box can be drawn.}
\end{inparaenum}

If the worker determines that at least one box can be drawn, the coreference chain proceeds to the Box Drawing task (below). Otherwise, we ask for a second and sometimes a third Box Requirement judgment to obtain agreement between two workers. If the majority agrees that no box needs to be drawn, the coreference chain is marked as ``non-visual" and leaves the bounding box annotation workflow. After preliminary analysis, we determined that coreference chains with mentions from the people, clothing, and body parts categories so frequently required boxes that they immediately proceeded to the Box Drawing task, skipping the Box Requirement task altogether.

\subsubsection{Box Drawing}
In this task, we collect bounding boxes for a mention. The key source of difficulty here is due to mentions that refer to multiple entities. Our annotation instructions specify that we expect individual boxes around each entity if these can be clearly identified (e.g., \textit{two people} would require two boxes). But if individual elements of a group cannot be distinguished (\textit{a crowd of people}), a single box may be drawn around the group. We show workers all previously drawn boxes for the representative mention (if they exist), and ask them to draw one new box around one entity referred to by the mention, or to indicate that no further boxes are required (see Figure~\ref{fig:interfaces}(d) for a screenshot). 

If the worker adds a box, the mention-box pair proceeds to the Box Quality task. If the worker indicates that no boxes are required, the mention accrues a ``no box needed" judgment.  The mention is then returned to Box Requirement if it has no boxes associated with it. Otherwise, the mention is sent to Box Coverage.

\subsubsection{Box Quality}
For each newly drawn box, we ask a worker whether the box is good. Since we want to avoid redundant boxes, we also show all previously drawn boxes for the same mention. Good boxes are tightly drawn around the entire entity a mention refers to which no other box already covers. When mentions refer to multiple entities that can be clearly distinguished, these must be associated with individual boxes. If the worker marks the box as Bad, it is discarded and the mention is returned to the Box Drawing task. If the worker marks the box as Good, the mention proceeds to the Box Coverage task to determine whether additional boxes are necessary. See Figure~\ref{fig:interfaces}(e) for an example interface for this task.

\subsubsection{Box Coverage}
In this step, workers are shown the boxes that have been drawn for a mention, and asked if all required boxes are present for that mention (Figure~\ref{fig:interfaces}(f)). If the initial judgment says that more boxes are needed, the mention is immediately sent back to Box Drawing. Otherwise, we require a second worker to verify the decision that all boxes have been drawn. If the second worker disagrees, we collect a third judgment to break the tie, and either send the mention back to Box Drawing, or assume all boxes have been drawn.

\subsection{Quality Control} 
\label{sec:quality}
Since worker quality on AMT is highly variable~\citep{sorokin2008utility,rashtchian2010collecting}, we take a combination of measures to ensure the integrity of annotations. First, we only allow workers who have completed at least 500 previous HITs with 95\%  accuracy, and have successfully completed a corresponding qualification test for each of our six tasks.  After this basic filtering, it is still necessary to ensure that a worker continues to provide quality annotations. A common method for doing so is to insert verification questions (questions with known answers) in all the jobs.  Initially, we included 20\% verification questions in our jobs, which were evaluated on a per-worker basis in batches.  While this process produced satisfactory results for the first three steps of the annotation pipeline (Binary Coreference Link Annotation, Coreference Chain Verification, and Box Requirement), we were not able to successfully apply this model to the last three steps having to do with box drawing. This appears to be due, in part, to the greater difficulty and attention to detail required in those steps.  Not only does someone have to read and understand the sentence and how it relates to the image being annotated, but he or she must also be careful about the placement of the boxes being drawn.  This increased difficulty led to a much smaller portion of workers successfully completing the tasks (see rejection rates in Table~\ref{table:annoStats}).  Even our attempts to change the qualification task to be more stringent had little effect on worker performance. Sticking with a verification model for these challenging tasks would either lead to higher costs (if we were to pay workers for poorly completed tasks) or greatly reduced completion rates (due to workers not wanting to risk doing a task they may not get paid for).

Instead, we used a list of Trusted Workers to pre-filter who can do our tasks.  To determine if a worker was to be placed on this list, those who passed our up-front screening were initially given jobs that only contained verification questions when they requested a job in our current batch.  If they performed well on their first 30 items based on the thresholds in Table~\ref{table:annoStats}, they would qualify as a Trusted Worker and would be given our regular jobs with only 2\% verification questions inserted.  To remain on the Trusted Worker list, one simply had to maintain the same quality level in both overall and most recent set of responses to verification questions.  The reduced number of verification questions limited the cost since poorly performing workers were identified quickly and more new items would be annotated for each job, which also increased the collection rate for our annotations.

\begin{table*}
\caption{Per-task crowdsourcing statistics for our annotation process.  Trusted Worker Quality is the average accuracy of trusted workers on verification questions (or approved annotations in the Box Drawing task). Min Performance is the Worker Quality score a worker must maintain to remain approved to do our tasks. To give an idea of the general level of complexity of our different tasks, we also list \% Rejected, which is the proportion of automatically rejected jobs (tasks) among {\em non-trusted} workers based on verification question performance. After we switched to a Trusted Worker model, we had virtually no rejected jobs.}
\label{table:annoStats}
\small
\begin{center}
\begin{tabular}{|l|r|r|r|r|r|r|}
\hline
& Annotations& \multirow{2}{*}{Avg Time (s)} & Num Trusted & Trusted Worker & \multirow{2}{*}{Min Performance} & \% Rejected For Non-\\
& Per Task & & Workers  & Quality & & Trusted Workers\\
\hline
Coreference Links & 10 & 75 & 587 & 90.6\%\textsuperscript{*} & 80\% & 2\textsuperscript{*}\\
Coreference Verify & 5 & 95 & 239 & 90.6\%\textsuperscript{*} & 83\% & 2\textsuperscript{*}\\
Box Requirement & 10 & 81 & 684 & 88.4\% & 83\% & $<$ 1\\
Box Drawing & 5 & 134 & 334 & 82.4\% & 70\% & 38.3\\
Box Quality & 10 & 110 & 347 & 88.0\% & 78\% & 52.7\\
Box Coverage & 10 & 91 & 624 & 89.2\% & 80\% & 35.4\\
\hline
\end{tabular}
\vspace{-1.5mm}
\flushleft\hspace{5mm}\begingroup
    \fontsize{6pt}{6pt}\selectfont
 *combined
\endgroup
\vspace{-3mm}
\end{center}
\end{table*}

\subsubsection{Additional Review}
At the end of the crowdsourcing process, we identified roughly 4k entities that required additional review. This included some chunking errors that came to our attention (e.g., through worker comments), as well as chains that cycled repeatedly through the Box Requirement or Box Coverage task, indicating disagreement among the workers. Images with the most serious errors were manually reviewed by the authors.

\begin{figure*}[t]
\centering
\topinset{\large\bfseries(a)}{\topinset{\large\bfseries(b)}{\includegraphics[width=\textwidth,trim=5.9cm 4.cm 0 0]{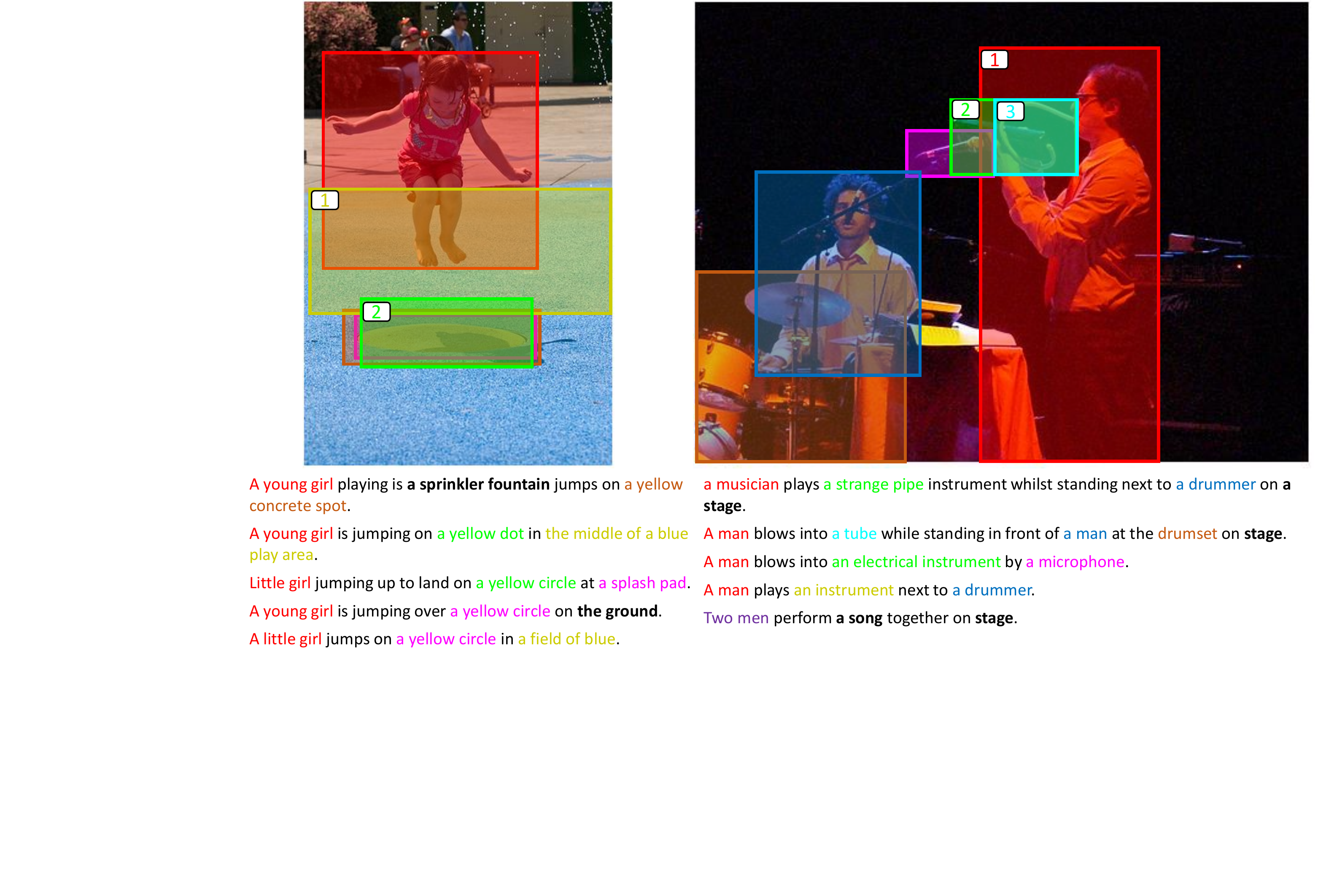}}{0in}{-0.75in}}{0in}{-3.3in}
\vspace{-15mm}
\caption{Examples of errors in Flickr30k Entities. In example (a), the second caption contains an error due to complex constructions. Here, the proper chunking should be {\em [the middle] of [a blue play area]}, where the {\em blue play area} is the entire blue region, and {\em the middle} refers to just the area containing the yellow dot. As it is, the coreference link {\em the middle of a blue play area} and {\em a field of blue} is not valid and there is an ambiguity as to whether the corresponding tan box (labeled 1) should cover just the yellow area or the entire blue area (either way, the box is incorrect). Furthermore, the entity mentions {\em a yellow dot}, {\em a yellow circle}, {\em a splash pad}, and {\em a yellow concrete spot} is fragmented into three chains with three distinct bounding boxes (labeled 2). 
In (b) the coreferent entity mentions {\em a strange pipe}, {\em a tube}, {\em an electrical instrument}, and {\em an instrument} are fragmented into three chains. 
The phrase \emph{an instrument} in the fourth sentence is linked to both boxes 1 and 2, when it should be linked to box 2 alone.  Box 3 for {\em a tube} is also too small, so it could not be merged with box 2.}
\label{fig:badExamples}
\end{figure*}

\subsubsection{Box and Coreference Chain Merging}
\label{sec:mergingprocess}
As discussed in Section \ref{sec:coref}, coreference chains may be fragmented due to missed links (false negative judgments).
Additionally, if an image contains more than one entity of the same type, its coreference chains may overlap or intersect (e.g., {\em a bride} and {\em a couple} from Figure \ref{fig:Examples}).
Since Box Drawing operates over coreference chains, it results in redundant boxes for such cases. We remove this redundancy by merging boxes with IOU scores of at least 0.8 (or 0.9 for ``other''). These thresholds were determined after an extensive manual review of the annotations. Some restrictions were placed on the types of phrases that were allowed to be combined (e.g. clothing and people boxes cannot be merged). Afterwards, we merge any coreference chains that point to the exact same set of boxes.  This merging resulting in a reduction of the number of bounding boxes in the dataset by 19.8\%  and 5.9\% fewer coreference chains.

\subsubsection{Error Analysis}

Errors present in our dataset mostly fall under two categories: chunking and coreference errors.  Chunking errors occur when the automated tools made a mistake when identifying mentions in caption text.  Coreference errors occur when AMT workers made a bad judgment when building coreference chains.  An analysis using a combination of automated tools and manual methods identified chunking errors in less than 1\% of the dataset's mentions and coreference errors in less than 1\% of the dataset’s chains. Since, on average, there are over 16 mentions and 7 chains per image, there is an error of some kind in around 8\% of our images. Figure~\ref{fig:badExamples} shows examples of some of the errors found in our dataset.

\subsection{Dataset Statistics} \label{sec:statistics}

\begin{figure}
\includegraphics[width=0.9\textwidth]{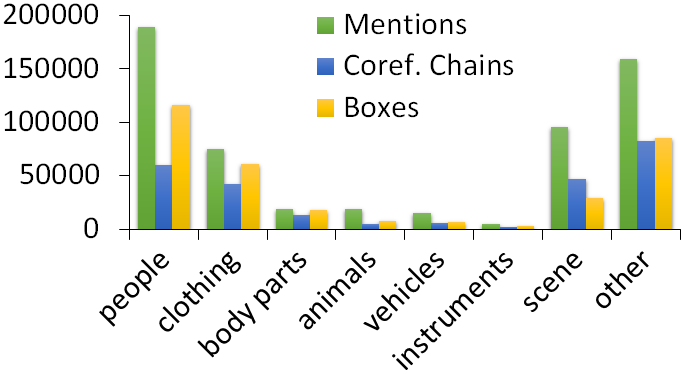}\vspace{-7pt}
\caption{The total number of coreference chains, mentions, and bounding boxes per type.}
\label{fig:Stats}
\end{figure}

\begin{table}
	\caption{Coreference chain statistics. The number of mentions per chain indicates how salient an entity is. The number of boxes per chain indicates how many distinct entities it refers to.}
    \label{table:entity_by_type}
\small
	\begin{tabular}{|l|r|c|c|c}
    	\hline
		Type & \#Chains & Mentions/Chain  &
        Boxes/Chain\\
		\hline
		people & 59,766 & 3.17 &	1.95\\
		clothing & 42,380 & 1.76 & 1.44\\
		body parts & 12,809 & 1.50 & 1.42\\
		animals	& 5,086 & 3.63 & 1.44\\
		vehicles & 5,561 & 2.77 & 1.21\\
		instruments	& 1,827 & 2.85 & 1.61\\
        scene & 46,919 & 2.03 & 0.62\\
		other & 82,098 & 1.94 & 1.04\\
        total & 244,035 & 2.10 & 1.13\\\hline
    \end{tabular}
\end{table}

Our annotation process has identified 513,644 entity or scene mentions in the 158,915 Flickr30k captions (3.2 per caption), and these have been linked into 244,035 coreference chains (7.7 per image). The box drawing process has yielded 275,775 bounding boxes in the 31,783 images (8.7 per image).
Figure \ref{fig:Stats} shows the distribution of coreference chains, mentions, and bounding boxes across types, and Table \ref{table:entity_by_type} shows additional coreference chain statistics.  48.6\% of the chains contain more than a single mention. The number of mentions per chain varies significantly across entity types, with salient entities such as people or animals being mentioned more frequently than clothing or body parts. 

Aggregating across all five captions, people are mentioned in 94.2\% of the images, animals in 12.0\%, clothing and body parts in 69.9\% and 28.0\%, vehicles and instruments in 13.8\% and 4.3\%, while other objects are mentioned in 91.8\% of the images. The scene is mentioned in 79.7\% of images. 59.1\% of the coreference chains are associated with a single bounding box, 20.0\% with multiple bounding boxes (with at least one such chain in 67.0\% of images), and 20.9\% with no bounding box, but there is again wide variety across entity types. The people category has significantly more boxes than chains (116k boxes for 60k chains) suggesting that many of these chains describe multiple individuals (\textit{a family}, \textit{a group of people}, etc.).  On average, each bounding box in our dataset has IOU of 0.37 with one other ground truth box and 49.2\% of boxes are completely enclosed by another ground truth box.

\begin{figure*}
\centering
\topinset{\large\bfseries(a)}{\includegraphics[width=0.9\textwidth]{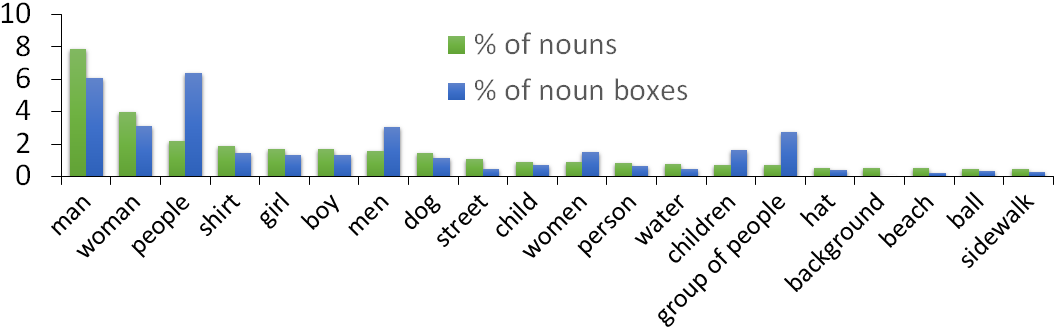}}{-.15in}{-3.4in}\vspace{-3mm}
\topinset{\large\bfseries(b)}{\includegraphics[width=0.9\textwidth]{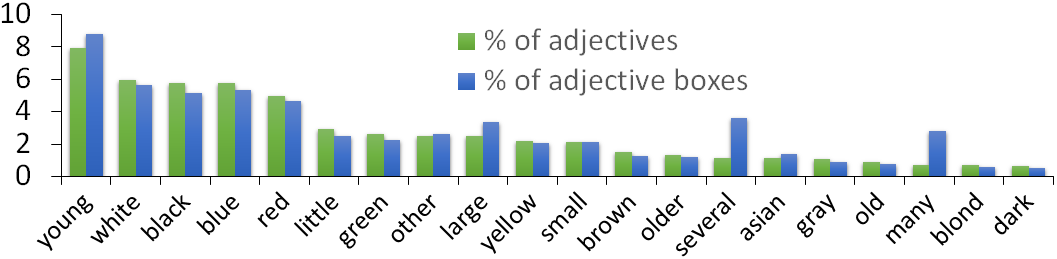}}{-.15in}{-3.4in}
\caption{Proportion of bounding boxes and occurrences across Flickr30k Entities of the most common {\bf (a)} nouns and {\bf (b)} adjectives.}
\label{fig:commonPhrases}
\end{figure*}

The 20 most common nouns and adjectives with their proportions of total boxes and occurrences are shown in Figure~\ref{fig:commonPhrases}.  Unsurprisingly, common nouns referring to people dominate, and adjectives referring to color appear quite often.  Some phrases that could be referring to a scene or a specific image region are also quite common (e.g.\ \emph{street, water}), providing a glimpse at the challenge faced when attempting to localize phrases since one would have to first identify the sense with which a phrase is being used.

%% file: phraseLocalization.tex
\section{Experimental Evaluation} 
\label{sec:results}

Our main motivation in collecting Flickr30k Entities is to further the development of methods that can reason about detailed correspondences between phrases in text and regions in an image. To evaluate this ability, we propose the following {\em phrase localization} benchmark: given an image and a ground-truth sentence that describes it, predict a bounding box (or bounding boxes) for each of the entity mentions (NP chunks) from that sentence. In Section \ref{sec:localization} we present a strong phrase localization baseline trained with our annotations, and in Section \ref{sec:retrieval}, we attempt to use it to improve performance on the standard task of bidirectional image-sentence retrieval.

\subsection{Phrase Localization} \label{sec:localization}

\subsubsection{Region-Phrase Model}
\label{sec:regionPhrase}

We have developed a baseline approach for phrase localization that scores each region-phrase correspondence separately, without taking into account any context or performing any joint inference about the global correspondence between all regions in the image and all phrases in the sentence. This approach learns an embedding of region and phrase features to a shared latent space and uses distance in that space to retrieve image regions given a phrase. While there have been several neural network-based approaches for learning such embeddings~\citep{karpathy2014deep,kiros2014unifying,mao2014deep}, using state-of-the-art text and image features with Canonical Correlation Analysis (CCA)~\citep{hotelling1936relations} continues to produce remarkable results~\citep{gong2014improving,klein2014fisher,klein2015rnn}, and is also much faster to train than a neural network. Given two sets of matching features from different views (in our case, image and text features), CCA finds linear projections of both views into a joint space of common dimensionality in which the correlation between the views is maximized.

Our implementation generally follows the details in~\cite{klein2014fisher}. Given a phrase, we represent each word with a 300-D word2vec feature~\citep{mikolov2013distributed} encoding only nouns, adjectives, and prepositions. Then we construct a Fisher Vector codebook~\citep{perronnin2010improving} with 30 centers using a Hybrid Gaussian-Laplacian Mixture Model (HGLMM),\footnote{Although in~\citet{klein2014fisher} their combined HGLMM+GMM Fisher Vectors performed the best on bidirectional retrieval, in our experiments the addition of the GMM features made no substantial impact on performance.} resulting in phrase features of dimensionality $300 \times 30\times 2=18,000$.   As in~\citet{klein2014fisher}, we report results using the 4096-dimensional activations of the 19-layer VGG model~\citep{simonyan2014very}, using a single crop of each ground truth region. We experiment with both classification and detection variants of the VGG network: the former is trained on the ImageNet dataset~\citep{deng2009imagenet} and the latter is the Fast RCNN network~\citep{girshickICCV15fastrcnn} fine-tuned on a union of the PASCAL 2007 and 2012 trainval sets~\citep{pascal-voc-2012}.  

An important implementation issue for training the CCA model is how to sample region-phrase correspondences from the training dataset. If we train CCA using all region-phrase correspondences, we get poor performance because the distribution of region counts for different NP chunks is very unbalanced: a few NP chunks, like {\em a man}, are extremely common, while others, like {\em tattooed, shirtless young man}, occur quite rarely. We found we can alleviate this problem by keeping at most $N$ randomly selected exemplars for each phrase, and we get our best results by resampling the dataset with $N=10$ regions per phrase. It is also important to note that in some images, a phrase can be associated with multiple regions (e.g., {\em two men}). In such cases, we merge the regions into a single bounding box for simplicity (although in follow-up work, it would be much more satisfying to detect the individual instances separately). 

Consistent with~\cite{klein2014fisher}, we set the CCA output dimensionality to 4096. To score region-phrase pairs using the learned CCA embedding, we use the normalized formulation of~\citet{Gong2014}, where we scale the columns of the CCA projection matrices by the eigenvalues and normalize feature vectors projected by these matrices to unit length. In the resulting space, we use cosine distance to rank image regions given a phrase.

\begin{table*}
\begin{center}
\caption{Overall phrase localization performance across the Flickr30k Entities test set. (a) Competing state-of-the-art methods. Note that these works use 100 Selective Search~\citep{selectivesearch} or EdgeBox proposals while we use 200 EdgeBox proposals. (b-d) Variants of our CCA model with different features or additional score terms added (see text for details).}
\label{table:phraseloc}
\begin{tabular}{|ll|c|c|c|}
\hline
& Methods & R@1 & R@5 & R@10 \\
\hline
\hline
 (a) & \citet{zhang2016EB} & { 27.0 } & { 49.9 } & { 57.7 } \\
 & \citet{hu2016natural} - VGG19  & { 27.8 } & { -- } & { 62.9 }  \\
 & \citet{rohrbach2015} - VGG19 & { 41.56 } & { -- } & { -- } \\
& \citet{wang2016matching} - Fast RCNN  & { 42.08 } & { -- } & { -- } \\
& \citet{deepspite2015} - Fast RCNN  & { 43.89 } & { 64.46 } & { 68.66 } \\
& \citet{rohrbach2015} - Fast RCNN  & { 47.70 } & { -- } & { -- }  \\
& \citet{fukui16emnlp} - Fast RCNN  & { 48.69 } & { -- } & { -- } \\
\hline
(b) & CCA - VGG19 & { 30.83 } & { 58.01 } & { 67.15 } \\
& CCA - Fast RCNN & { 41.77 } & { 64.52 } & { 70.77 } \\
\hline
(c) & CCA or Detector & { 42.58 } & { 65.26 } & { 71.28 } \\
& CCA+Detector  & { 43.84 } & { 65.83 } & { 71.75 } \\
\hline
(d) & CCA+Detector+Size & { 49.22 } & { 69.93 } & { 74.90 } \\
& CCA+Detector+Color & { 45.79 } & { 67.23 } & { 72.86 } \\
& CCA+Detector+Size+Color & \bf{ 50.89 } & \bf{ 71.09 } & {\bf 75.73} \\
\hline
\end{tabular}
\end{center}
\end{table*}

\subsubsection{Evaluation Protocol} 
\label{sec:localeval}

At test time we assume we are given an image and a set of NP chunks from all of its ground truth captions.\footnote{We use ground truth NP chunks and ignore the non-visual mentions (i.e., mentions not associated with a box). The alternative evaluation method is to extract the phrases automatically, which introduces chunking errors and lowers our recall by around 3\%. To the best of our knowledge, the competing methods in Table~\ref{table:phraseloc}(a) also evaluate using ground-truth NP chunks.} We use the EdgeBox region proposal method~\citep{ZitnickECCV14} to extract a set of candidate object regions from the test image. Experimentally, we found 200 proposals to give us the best performance. Then, for each phrase, we rank the proposal regions using the CCA model and perform non-maximum suppression using a 0.5 IOU threshold.

Following~\citet{gong2014improving,karpathy2014deep,klein2014fisher,mao2014deep}, we split Flickr30K into 29,783 training, 1,000 validation, and 1,000 test images. Our split is the same as in~\citet{gong2014improving}. We evaluate localization performance by treating the phrase as the query to retrieve the proposals from the input image and report Recall@$K$ ($K = 1,5,10$), or the percentage of queries for which a correct match has rank of at most $K$ (we deem a region to be a correct match if it has IOU $\ge$ 0.5 with the ground truth bounding box for that phrase). 

Note that in the initial version of this work~\citep{flickrentities}, we reported average precision (AP) numbers in addition to recall. However, our annotations are very sparse: there are many valid regions corresponding to some phrases, especially body parts and clothing, that lack ground truth bounding boxes because they are never mentioned in captions. This pervasive reporting bias for some phrase types, combined with the rarity of other phrase types, makes AP too unreliable. Thus, consistent with other works that perform evaluation on Flickr30K Entities~\citep{fukui16emnlp,hu2016natural,rohrbach2015,wang2016matching}, we only report recall in this paper.


\begin{figure*}
\centering
\includegraphics[height=6cm]{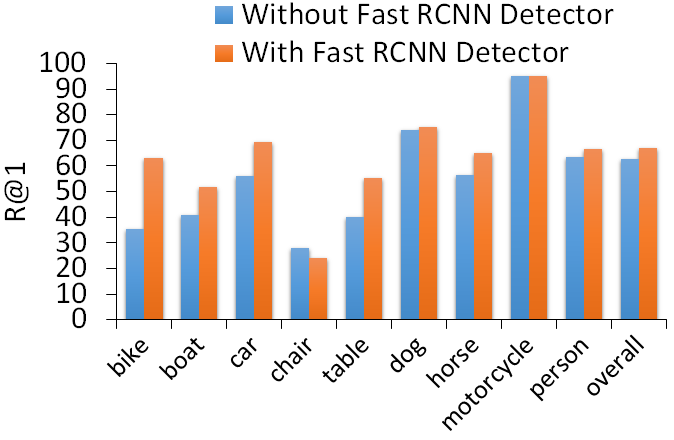}
\caption{Comparison over PASCAL object categories that occur at least 20 times in the test set showing how averaging the CCA score with the output of the Fast RCNN detector affects phrase localization performance.
\label{fig:pascalObjects}}
\end{figure*}

\begin{figure*}
\centering
\topinset{\large\bfseries(a)}{\includegraphics[height=6cm,trim=0 2.2cm 2cm 0]{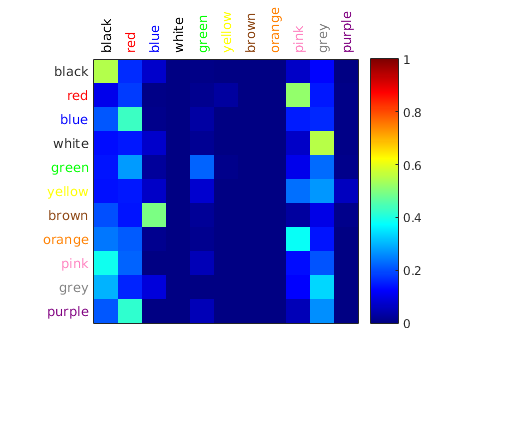}}{0.2in}{-1.2in}
\hspace{0.4cm}\topinset{\large\bfseries(b)}{\includegraphics[height=6cm,trim=2 2.2cm 2cm 0]{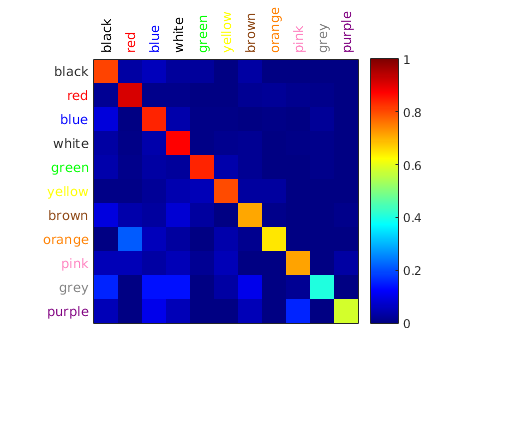}}{0.2in}{-1.2in}
\vspace{-3mm}
\caption{Confusion matrices for color classification on the test set using {\bf (a)} linear SVM trained on fc7 features computed from a Fast RCNN network fine-tuned on PASCAL object classes or {\bf (b)} a Fast RCNN network trained to predict colors. Colors are ordered from most to least prevalent in the dataset.}
\label{fig:colorConfusion}
\end{figure*}

\subsubsection{Phrase Localization Experiments}

Table~\ref{table:phraseloc} summarizes the results of our phrase localization experiments. For reference, part (a) of the table lists recent results on this task which generally fall under two categories: LSTM-based methods~\citep{hu2016natural,rohrbach2015,fukui16emnlp} and those that use a shallow neural network to learn an embedding between text and image features~\citep{deepspite2015,wang2016matching}.
We also include the performance of the neural attention model of \citet{zhang2016EB}, but note that it is a weakly supervised method trained on outside data. 

From Table \ref{table:phraseloc}(b), we can see that switching from the classification-based VGG19 network, which was used in the initial version of our work~\citep{flickrentities}, to the detection-based Fast RCNN network improves accuracy significantly, which is consistent with the observations of \citet{rohrbach2015}. However, the localization quality of CCA is fundamentally limited because it is trained only on positive examples (ground-truth regions and corresponding phrases). Ideally, we would prefer to use an actual detector that is also trained using negative examples, i.e., poorly localized and background regions. On the other hand, by using a continuous text embedding, CCA can better cope with rare and unseen phrases, as well as phrases that are semantically related. To combine the advantages of both models, we put together the following hybrid scheme. 

We manually created mappings from subsets of phrases in our dataset to the 20 PASCAL object categories. These mappings affect 25.32\% of all our phrases, 83.4\% of which are from the ``person'' type. When we encounter one of these phrases at test time, we score the proposal regions using the full detection machinery of \cite{girshickICCV15fastrcnn}, including bounding box regression. We then get a combined score for phrase $\phi$ and region $r$ by averaging the detector and CCA scores:
\begin{equation}
D_{CCA+det}(\phi,r) = 0.5\,D_{CCA}(\phi,r) + 0.5(1-\sigma_{det}(\phi,r)) \,,
\end{equation}
where $D_{CCA}$ is the cosine CCA distance (which is between 0 and 1), and $\sigma_{det}$ is the softmax detector score (which is also between 0 and 1).
For phrases that do not correspond to a pre-trained detector, we use only the CCA score. As can be seen from Table \ref{table:phraseloc}(c), using the detector score alone for phrases that have it is better than using the CCA score alone, and using a combination of both works the best. Figure~\ref{fig:pascalObjects} compares the performance of CCA-only with the combined score over PASCAL categories that occur at least 20 times in our test set.

\begin{table*}[t]
\begin{center}
\begin{tabular}{|ll|c|c|c|c|c|c|c|c|}
\hline
& & people &  clothing & bodyparts & animals & vehicles  & instruments & scene & other \\
\hline
\hline
& \#Instances & { 5,656 }  & { 2,306 } & { 523 } & { 518 } & { 400 }  & { 162 } & { 1,619 } & { 3,374 }\\
\hline
R@1 & CCA+Detector & { 61.24 } & { 36.90 } & { 15.30} & { 62.74 } & { 59.75 } & { 31.48 } & { 31.93 } & { 25.34 }\\
& CCA+Detector+Size & { 64.73 } & { 39.20 } & { 15.49 } & { 64.09 } & { 67.75 } & { 37.65 } & { 51.33 } & { 30.50 }\\
& CCA+Detector+Size+Color & { 64.73 } & { 46.88 } & { 17.21 } & { 65.83 } & { 68.75 } & { 37.65 } & { 51.39 } & { 31.77 }\\
\hline
& Proposal upper bound (R@200) & { 96.52 } & { 77.36 } & { 50.48 } & { 91.12 } & { 94.50 } & { 80.86 } & { 83.01 } & { 75.87 }\\
\hline
\end{tabular}
\caption{Localization performance over phrase types to rank 200 object proposals per image.}
\label{table:phraseLocBreakdown}
\end{center}
\end{table*}

Next, we introduce two more additions to our CCA+Detector model to make it a very strong baseline indeed, rivaling the more complex method of \cite{rohrbach2015}. First, we observe that we can get a big improvement by introducing a bias towards larger regions. In fact, simply selecting the largest proposal regardless of the phrase already gets R@1 of 24\%. To trade off the appearance-based score with the region size, we define the following combined score:
\begin{eqnarray}
\lefteqn{D_{CCA+det+size}(\phi,r) = } \\ \label{eq:sizetradeoff}
& & (1-w_{size})D_{CCA+det}(\phi,r) + w_{size}(1-size(r)) \,, \nonumber
\end{eqnarray}

\noindent where $size(r)$ is the proportion of the image area the region $r$ covers.  The weight $w_{size}$ is separately determined for each of our eight phrase types based on the validation set (it is 0.2 for scene, vehicle, and instrument types, and 0.1 for everything else). The first line of Table~\ref{table:phraseloc}(d) shows this simple method works remarkably well, increasing R@1 and mAP by about six points.


Color can also be a strong indicator of the location of a phrase in an image, especially for clothing. However, image features fine-tuned for object detection, where objects of different colors may fall under the same category, turn out to be relatively insensitive to color. Specifically, if we train an SVM classifier on top of Fast RCNN features to predict one of eleven colors that occur at least 1,000 times in the training dataset, we get only 16\% accuracy (see Figure~\ref{fig:colorConfusion}(a)). 
To obtain a better color predictor for bounding boxes, we fine-tuned the Fast RCNN network on these eleven colors. To avoid confusion with color terms that refer to race, we excluded people phrases from training and testing. We used a softmax loss (i.e., color classification is assumed to be one-vs-all) and fine-tuned the whole network with 0.001 learning rate, 0.0005 weight decay, and 0.9 momentum for 20K iterations.  As can be seen in Figure~\ref{fig:colorConfusion}(b), the resulting network has a much higher accuracy of 80.47\%.  

With our new color classifier, we add a color term to eq. (2) to obtain our full model:
\begin{eqnarray} 
\lefteqn{D_{full}(\phi,r) = } \\ \label{eq:full} \nonumber 
& & (1-w_{size}-w_{color})D_{CCA+det}(\phi,r) + \\ \nonumber & & w_{size}(1-size(r)) + w_{color}(1-\sigma_{color}(\phi,r))\,, \nonumber
\end{eqnarray}
where $\sigma_{color}(\phi,r)$ is the softmax output of the classifier for the color mentioned in phrase $\phi$. We use this term for phrases that mention a color,\footnote{If a phrase includes more than one color, all the color mentions are ignored.} and eq. (2) otherwise. As can be seen from Table \ref{table:phraseLocBreakdown}, the resulting CCA+Size+Color model mainly improves the accuracy for the clothing phrase type, but because this type is so common, this leads to an approximately 1.5\% improvement on the entire test set (last two lines of Table~\ref{table:phraseloc}(d)).

\begin{figure*}[h]
\centering
\topinset{\large\bfseries(a)}{\includegraphics[height=5cm]{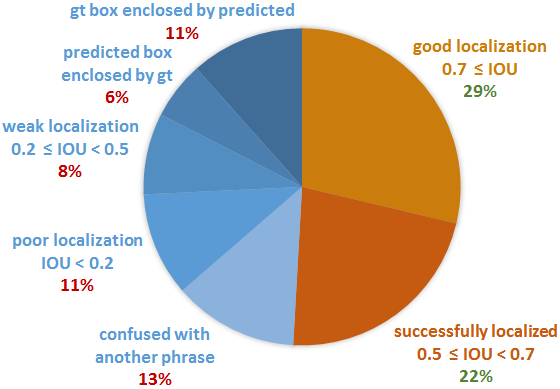}}{-0.18in}{-1.5in}
\hspace{2cm}\topinset{\large\bfseries(b)}{\includegraphics[height=5.3cm,trim=2cm 0cm 0cm 0.5cm]{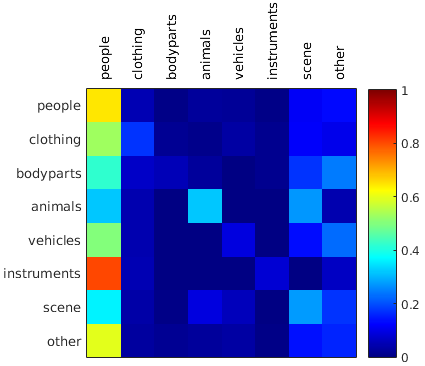}}{-.05in}{-1.5in}
\caption{{\bf (a)} A breakdown of the R@1 localization performance of our full model.  {\bf (b)} Confusion matrix for the 13\% of phrases that get confused with another phrase. The entry in row $i$ and column $j$ shows how often a phrase of type $i$ is localized to a box corresponding to phrase of type $j$. For example, how often does a poorly localized bounding box for a phrase of type ``clothing'' have $\geq$ 0.5 IOU with the ground truth box for a phrase of type ``people''? The matrix calls attention to a pattern of predicting a bounding box for a person when the model is unsure about the location of a phrase.}
\label{fig:phraseLocBreakdown}
\end{figure*}

\subsubsection{Phrase Localization Discussion}
\label{sec:localdicussion}

As can be seen in the last line of Table~\ref{table:phraseloc}(d), our full model performs relatively well, accurately localizing a phrase more than 50\% of the time in an image that contains that phrase. Table \ref{table:phraseLocBreakdown} shows a detailed breakdown that gives an idea of how our different cues contribute to the performance on different phrase types, and the relative difficulty of these phrase types. We can see that adding the size term gives the biggest improvement for vehicles and scenes. For phrases from the scene type, we also experimented with simply predicting the whole image, but that did not give better performance, possibly due to the ambiguity of some phrases (in some cases, {\em building} may refer to the whole image, and in some cases, it may refer to an object that occupies just a part of the image). As mentioned above, the color term gives the biggest improvement for clothing. It also helps with the body parts mainly due to improved ability to detect hair based on color (brown, black, gray, and even blue or pink).

In absolute terms, we get by far the lowest accuracy on body parts, followed by clothing and instruments (though the latter have just a few instances). This difficulty is due at least in part by the poor coverage that our region proposals give for these classes -- as can be seen from the ``Proposal upper bound'' line of Table \ref{table:phraseLocBreakdown}, only about 50\% of body parts and 77\% of clothing items have a box in our entire set of 200 region proposals with at least 50\% IoU. We found that simply adding more region proposals decreased the precision for these phrase types, so their complex appearance adds to the challenge as well.

\begin{figure*}
\centering
\includegraphics[height=4cm]{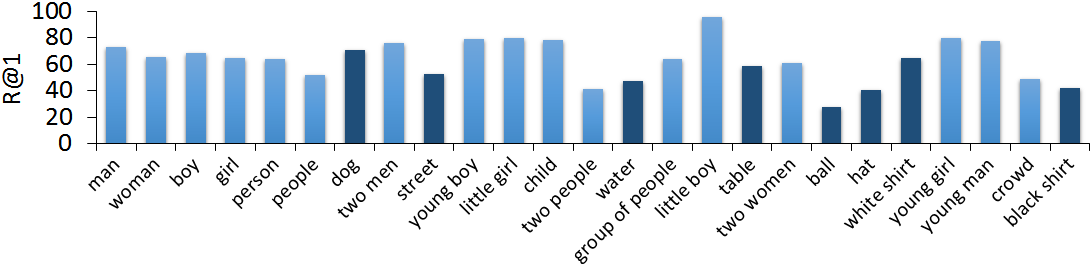}
\caption{Localization performance of 25 of the most common phrases in the test set using our full model ranking 200 object proposals per image. Darker color indicates phrases that are not from the people type.}
\label{fig:wordLoc}
\end{figure*}

Figure~\ref{fig:phraseLocBreakdown}(a) analyzes the sources of errors our model makes, showing that confusion between phrases is one of the biggest sources. Figure~\ref{fig:phraseLocBreakdown}(b) shows a confusion matrix between different phrase types, revealing a bias towards predicting bounding boxes for a person. Figure~\ref{fig:wordLoc} shows the accuracies for the 25 most frequent phrases in our test set.

\begin{figure*}
\centering
\topinset{\large\bfseries(a)}{\includegraphics[width=0.95\textwidth,trim=0cm 3.5cm 0cm 0cm]{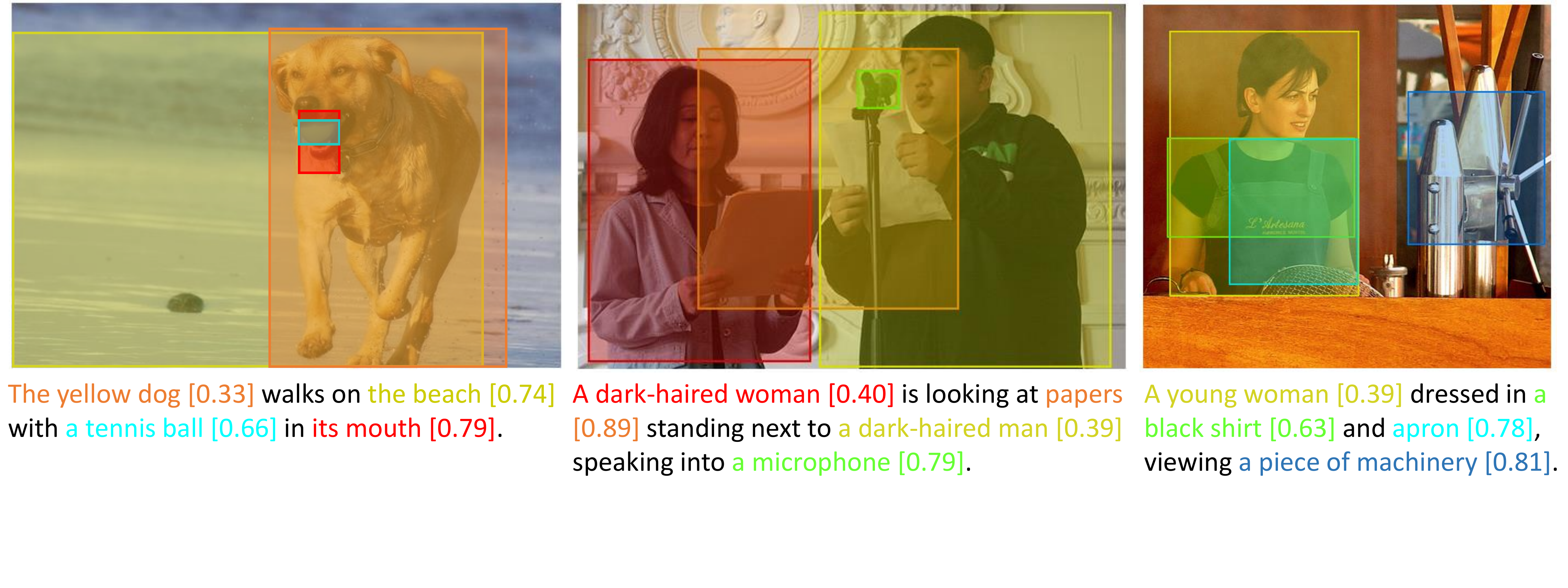}}{-.15in}{-3.3in}
\topinset{\large\bfseries(b)}{\includegraphics[width=0.94\textwidth,trim=0cm 5.5cm 0.1cm 0cm]{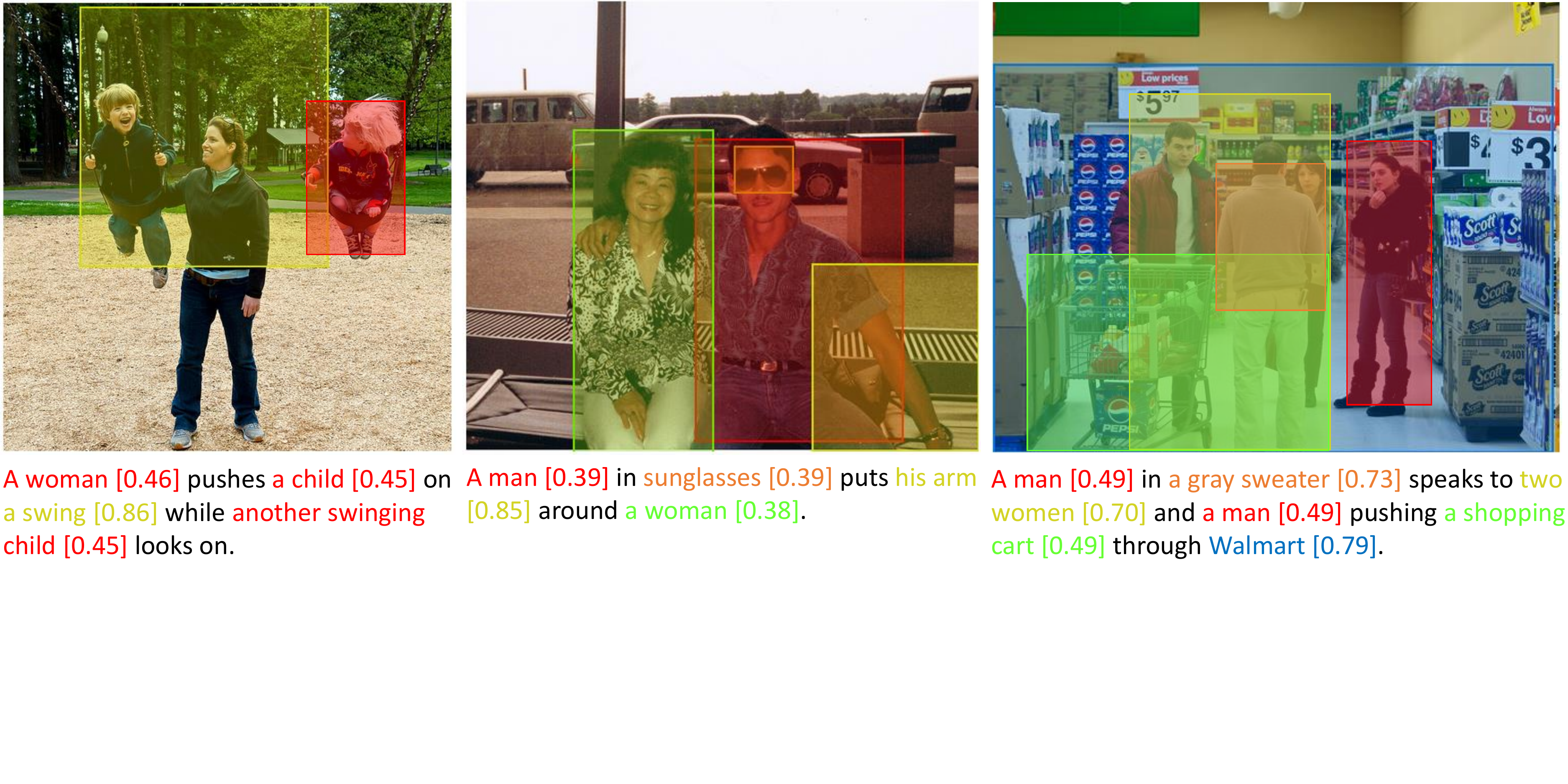}}{-.15in}{-3.3in}
	\caption{Example phrase localization results. For each image and reference sentence, phrases and top matching regions are shown in the same color. The matching score is given in brackets after each phrase (low scores are better). }
    \label{fig:phraseLocalizationOutput}
\end{figure*}

Figure~\ref{fig:phraseLocalizationOutput}(a) shows examples of relatively successful localization in three images. Our model can find small objects (e.g.\ \emph{a tennis ball} in the left example and \emph{a microphone} in the middle). In the middle example, it can correctly distinguish the man from the woman. Three typical failure modes are shown in Figure~\ref{fig:phraseLocalizationOutput}(b), reflecting our difficulties with localizing body parts and correctly disambiguating person instances. In the leftmost example, three different people phrases are localized to the same box. In the middle example, the bounding box for {\em arm} localizes the man's visible left arm, instead of the mentioned but mostly occluded {\em arm around a woman}. In the right example, there are several revealing errors. The bounding box for {\em two women}, while enclosing multiple people, is incorrect. Further, the boxes for two separate instances of {\em man} incorrectly land on the same woman even though the {\em gray sweater} belonging to one of the men is correctly localized. This is not surprising, since our model uses the phrase itself without any surrounding sentence context, so multiple instances whose mentions are identical must necessarily be localized to the same box; there is also no constraint in our model to enforce co-location of people and clothing or body parts. 



In order to go beyond our baseline, it is necessary to develop methods that can decode the textual cues about cardinalities of entities and relationships between them, and translate these cues into constraints on the localized regions. In particular, since people are so important for our dataset and for image description in general, it is necessary to parse a sentence to determine how many distinct persons are in an image, which mentions of clothes and body parts belong to which person, and impose appropriate constraints on the respective bounding boxes. This is subject of our future work.


%% file: retrieval.tex
\begin{table*}[t]
\centering
\caption{Bidirectional retrieval results. Image Annotation refers to using images to retrieve sentences, and Image Search refers to using sentences to retrieve images. The numbers in (a) come from published papers, and the numbers in (b) are from our own reproduction of the results of~\cite{klein2014fisher} using their code. See Section \ref{sec:retrieval} for additional details.}
\label{table:flickr30k}
\small
\begin{center}
\begin{tabular}{|l|l|l|l|l|l|l|l|}
\hline
& Methods on Flickr30k                         & \multicolumn{3}{l|}{Image Annotation}                                                                                     & \multicolumn{3}{l|}{Image Search}                                                                                        \\ \hline
&                                             & R@1                                    & R@5                                    & R@10                                   & R@1                                    & R@5                                    & R@10                                   \\ \hline
(a) State of the art  
& {{\citet{klein2014fisher}}}   & { {33.3\%}} & { {62.0\%}} & { {74.7\%}} & { {25.6\%}} & { {53.2\%}} & { {66.8\%}} \\
& {{\citet{mao2014deep}}}   & { {35.4\%}} & { {63.8\%}} & { {73.7\%}} & { {22.8\%}} & { {50.7\%}} & { {63.1\%}} \\ 
& {{\citet{ma2015}}}   & { {33.6\%}} & { \textbf {64.1\%}} & { {74.9\%}} & { {26.2\%}} & { \textbf{56.3\%}} & { {69.6\%}} \\ 
& {{\citet{klein2015rnn}}}   & { {35.6\%}} & { {62.5\%}} & { {74.2\%}} & { \textbf{27.4\%}} & { {55.9\%}} & { \textbf{70.0\%}} \\ 
\hline
(b) Whole image-sentence CCA 
& {{HGLMM FV+VGG19}}   & { {36.5\%}} & { {62.2\%}} & { {73.3\%}} & { {24.7\%}} & { {53.4\%}} & { {66.8\%}} \\
\hline


(c) Combined image-sentence & {{Weighted Distance VGG19}}  & { 37.0\% }       & { 62.9\%}        & { 73.9\%}      & { {25.7\%}}      & { {54.5\%}}    & { {67.6\%}} \\ 
and region-phrase & {{Weighted Distance Full Model}}  & { \textbf{37.5\%}}       & { 62.9\%}        & { \textbf{75.1\%}}      & { 25.8\%}      & { 54.7\%}    & { 67.6\%} \\ 
\hline
\end{tabular}
\end{center}
\end{table*}

\subsection{Image-Sentence Retrieval}
\label{sec:retrieval}
Next, we would like to demonstrate the usefulness of phrase localization for the well-established benchmark of bidirectional image-sentence retrieval: given an image, retrieve the best-fitting sentence from a pre-existing database, and vice versa. For this, we will start with a state-of-the-art CCA model trained on whole images and sentences, which already does a very good job of capturing the global content of the two modalities, and attempt to refine it using the region-phrase model of Section \ref{sec:regionPhrase}. Here, the region-phrase model has to succeed at a more difficult task than in Section \ref{sec:localization}: instead of scoring regions in an image to localize a phrase that is assumed to be present, it has to compare scores for different region-phrase combinations in an attempt to determine which combination provides the best description of the image.

To get the best global image representation, we use the ImageNet-trained 19-layer VGG network and average the whole-image features over ten crops. Apart from this, we follow the implementation details of Section~\ref{sec:regionPhrase} to train an image-sentence CCA model that is essentially a reimplementation of~\cite{klein2014fisher}. Given the model, we compute the normalized projections of the image and sentence features into the CCA space and do image-to-sentence and sentence-to-image retrieval using the cosine distance.

For evaluation, we use the standard protocol for Flickr30k: given the 1,000 images and 5,000 corresponding sentences in the test set, we use the images to retrieve the sentences and vice versa, and report performance as Recall@K, or the percentage of queries for which at least one correct ground truth match was ranked among the top $K$ matches. Table~\ref{table:flickr30k} shows the results. As can be seen by comparing Table \ref{table:flickr30k}(a) and (b), the global CCA has consistent performance with~\cite{klein2014fisher} and is competitive with the state of the art, which includes complex CNN and RNN models.

Next, we want to add region-phrase correspondences to get a further improvement on image-sentence matching. Given an image $I$ and a sentence $S$ (which may or may not correctly describe the image), for each phrase $\phi_{i}$, $i=1,\ldots,L$, we find the best-matching candidate region $r_j$ using the region-phrase CCA embedding.\footnote{Here, as in Section~\ref{sec:results}, our phrases are ground-truth NP chunks, but unlike in Section~\ref{sec:results}, we do not exclude NP chunks corresponding to non-visual concepts.} Then, similarly to~\cite{karpathy2014deep}, we compute the overall image-sentence distance as the sum of the region-phrase distances:

\begin{equation}\label{dis_ave}
  D_{PR}(S,I) = \frac{1}{L^\gamma} \sum_{i}^L \min_j D_{full}(\phi_i,r_j) \, ,
\end{equation}
where $D_{full}$ is our full region-phrase model (eq. 3) and the exponent $\gamma \ge 1$ is meant to lessen the penalty associated with matching images to sentences with a larger number of phrases, since such sentences tend to mention more details that are harder to localize. Experimentally, we have found $\gamma = 1.5$ to produce the best results. Finally, we define a combined image-sentence distance as 
\begin{equation}\label{dis}
D_{SI} = \alpha\, {D}_{CCA}(S,I) +  (1-\alpha)\,D_{PR}(S,I) \,,
\end{equation}
where $D_{CCA}(S,I)$ is the normalized CCA distance between the whole-image and whole-sentence feature vectors.

\begin{figure*}
\centering
\begin{tabular}{rc|c}
& \large Top Sentence From Whole Image-Sentence &  \large Top Sentence With Region-Phrase\\
& \large  &  \large \\
\large\bfseries(a) & \multicolumn{1}{l|}{ Image-Sentence Score: 0.54}  & \multicolumn{1}{l}{Image-Sentence Score: 0.58} \\
 &\multicolumn{1}{l|}{Region-Phrase Score: 0.49}  & \multicolumn{1}{l}{Region-Phrase Score: 0.33} \\
&\includegraphics[width=0.45\textwidth,trim=0.2cm 1cm 0cm 0cm]{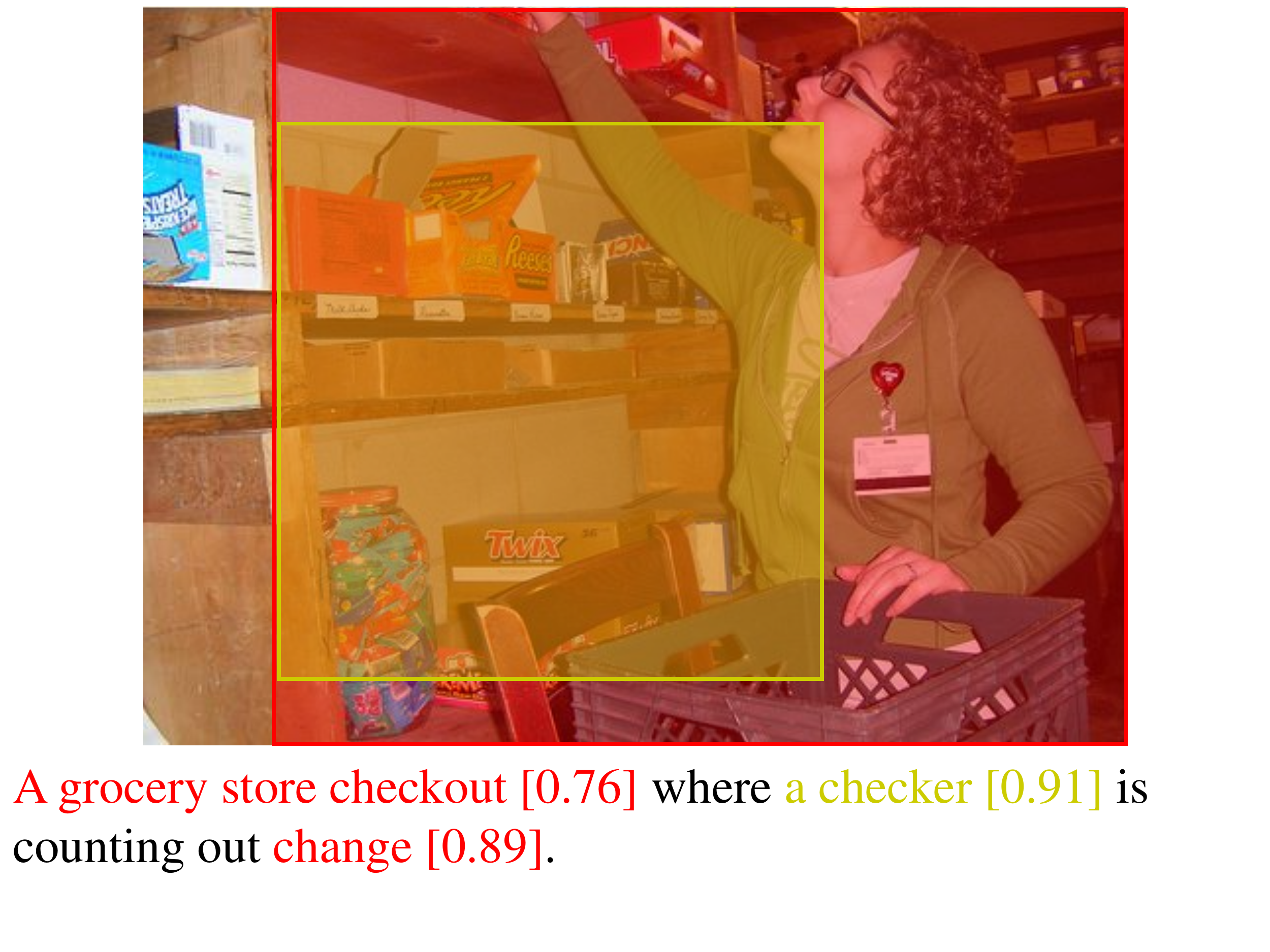} & \includegraphics[width=0.45\textwidth,trim=0.2cm 1cm 0cm 0cm]{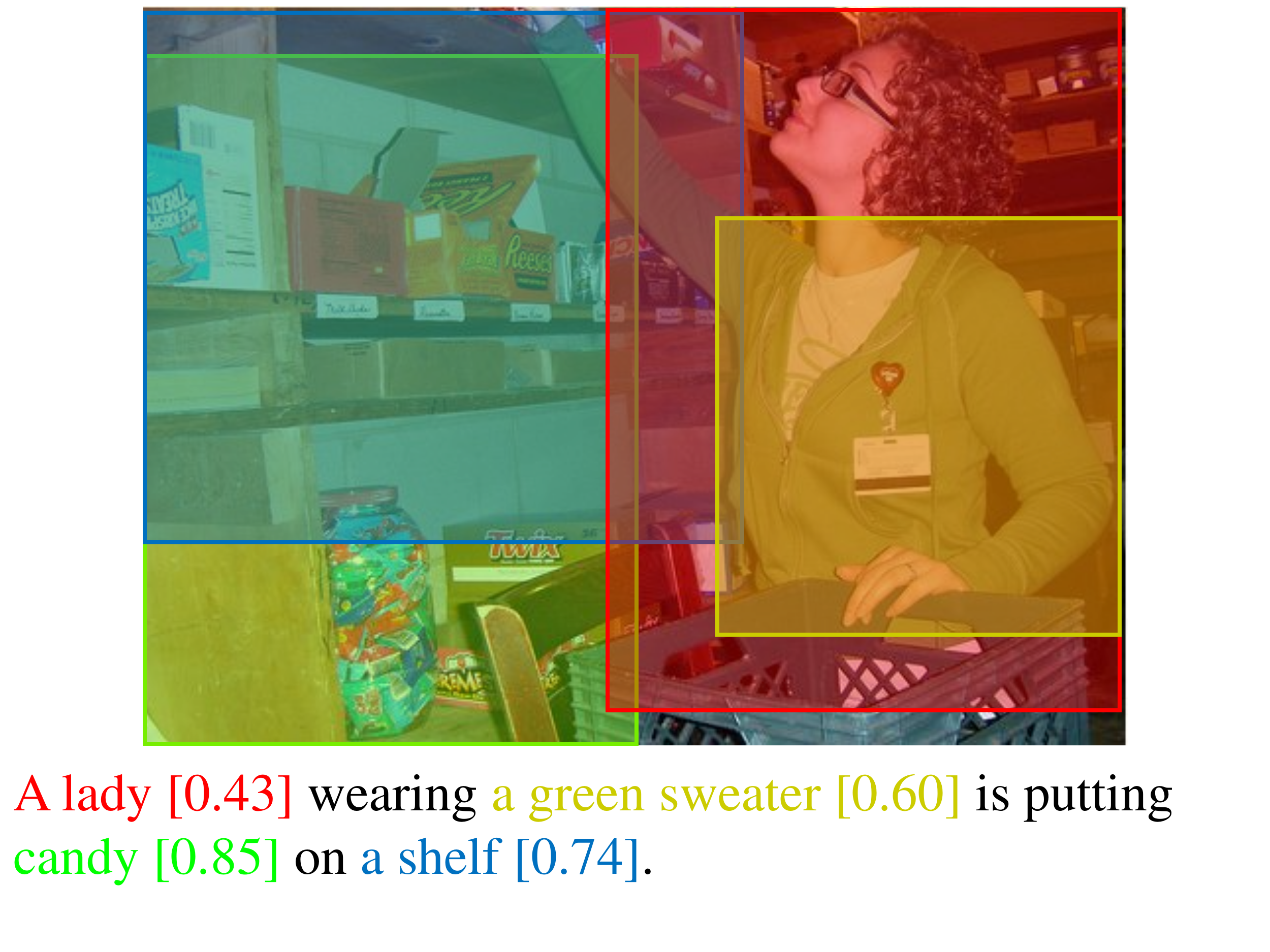} \\
\large\bfseries(b) & \multicolumn{1}{l|}{ Image-Sentence Score: 0.22}  & \multicolumn{1}{l}{Image-Sentence Score: 0.23} \\
 &\multicolumn{1}{l|}{Region-Phrase Score: 0.36}  & \multicolumn{1}{l}{Region-Phrase Score: 0.25} \\
&\includegraphics[width=0.45\textwidth,trim=0.2cm 1cm 0cm 0cm]{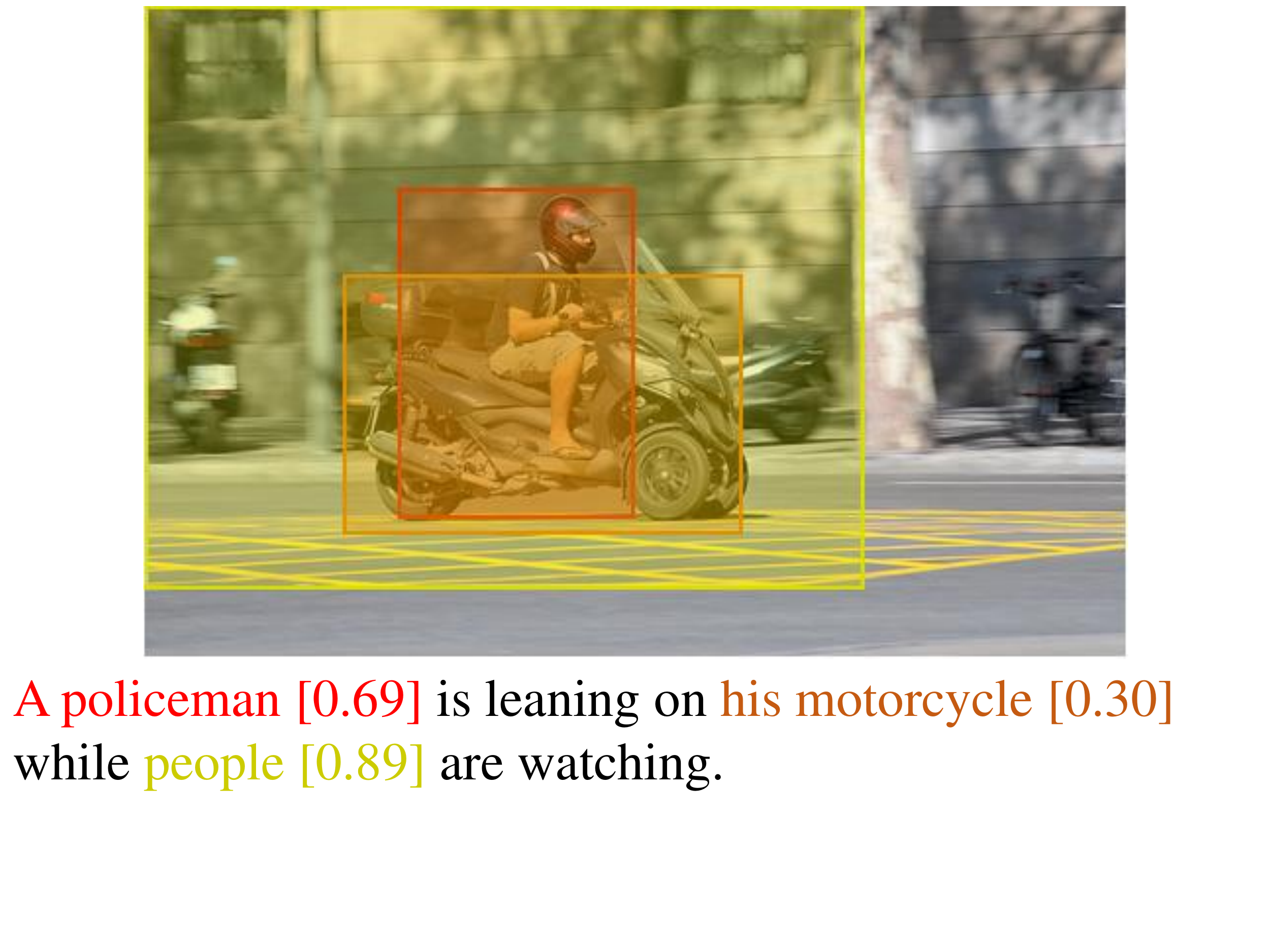} & \includegraphics[width=0.45\textwidth,trim=0.2cm 1cm 0cm 0cm]{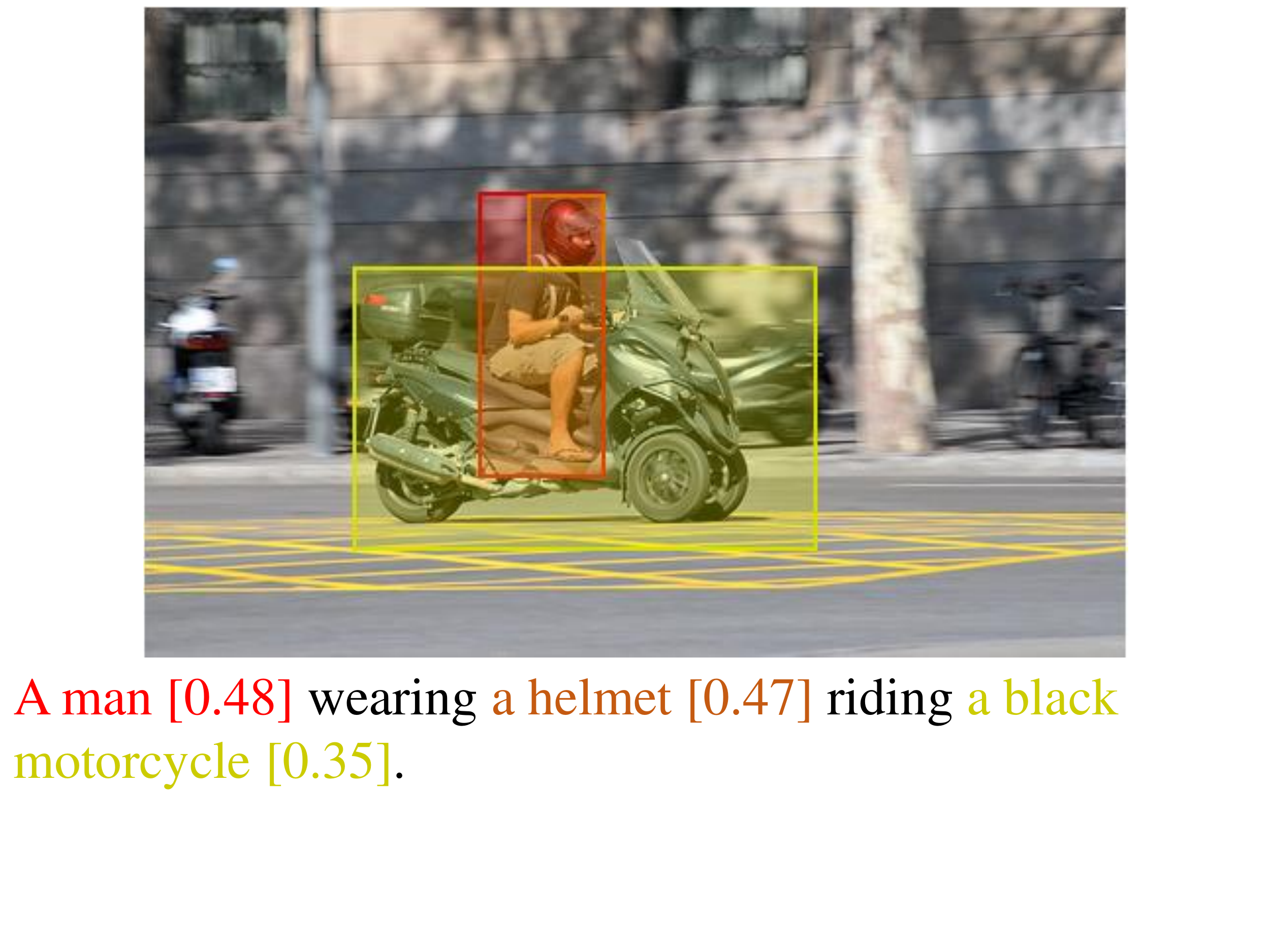} \\
\large\bfseries(c) & \multicolumn{1}{l|}{ Image-Sentence Score: 0.64}  & \multicolumn{1}{l}{Image-Sentence Score: 0.66} \\
 &\multicolumn{1}{l|}{Region-Phrase Score: 0.35}  & \multicolumn{1}{l}{Region-Phrase Score: 0.30} \\
&\includegraphics[width=0.45\textwidth,trim=0.2cm 1cm 0cm 0cm]{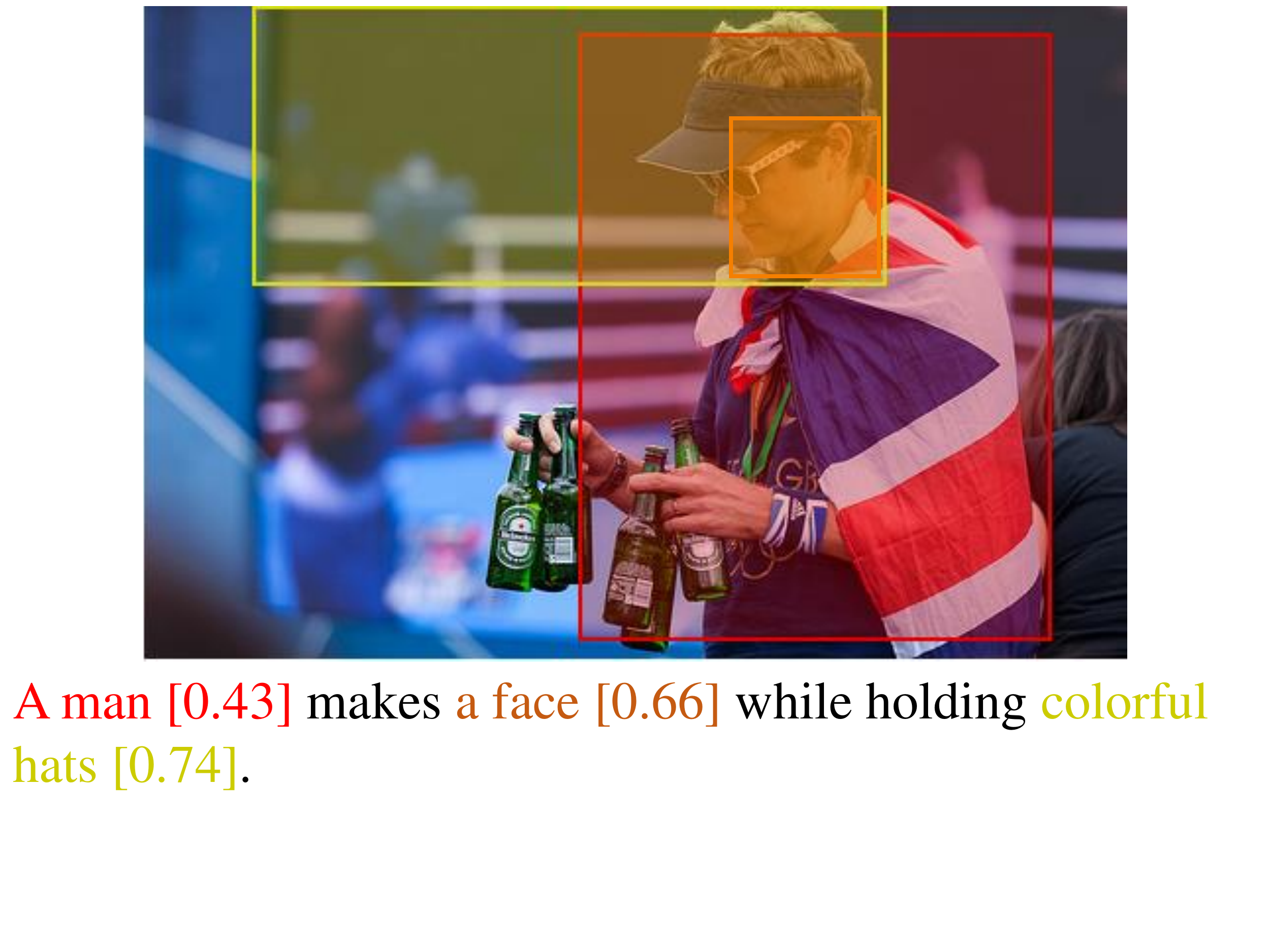} & \includegraphics[width=0.45\textwidth,trim=0.2cm 1cm 0cm 0cm]{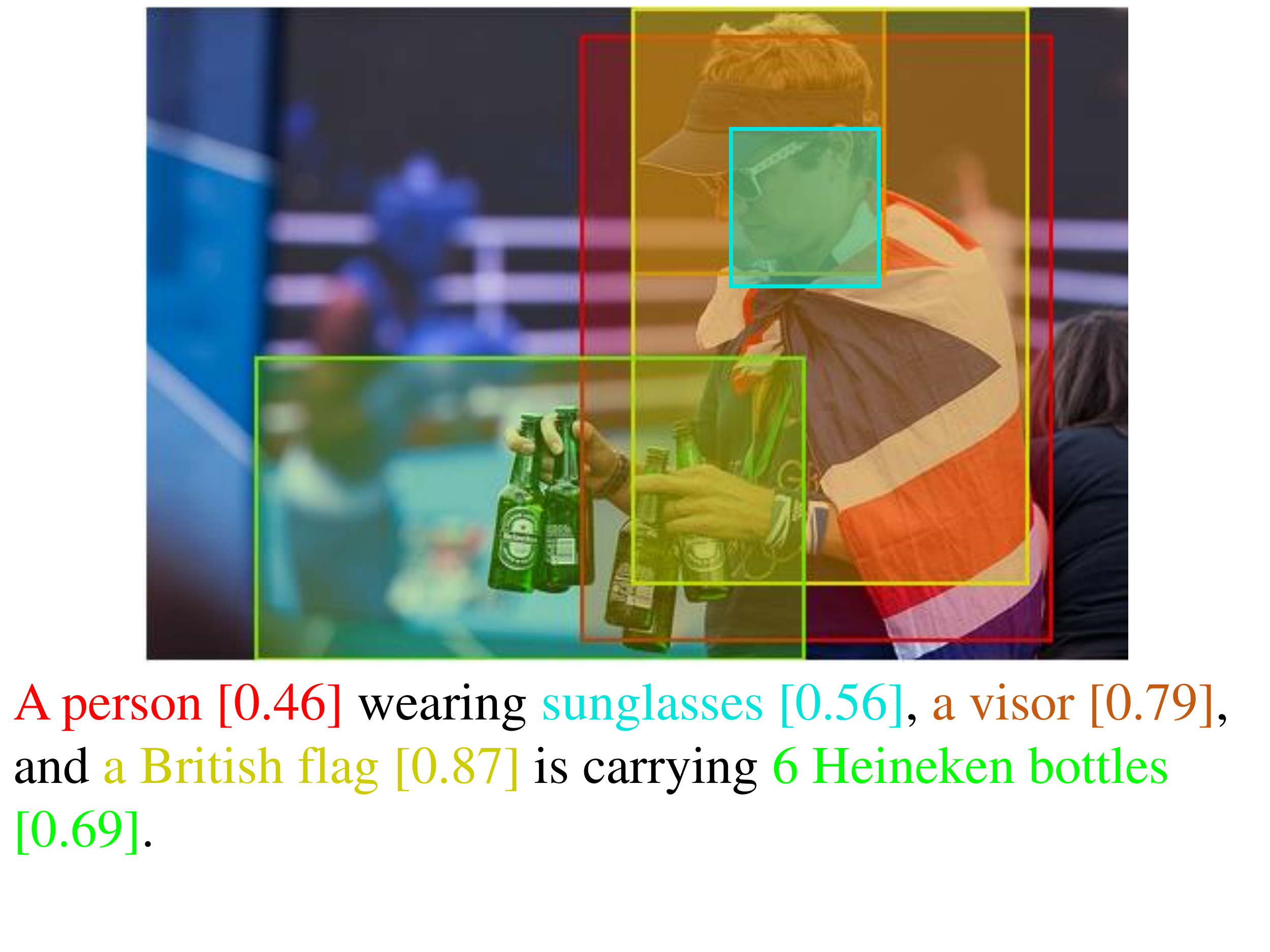} \\
\end{tabular}
	\caption{Example image-sentence retrieval results where adding region-phrase correspondences helps to retrieve the correct sentence.  For each test image, the left column shows the top retrieved sentence using the whole image-sentence model and the right column shows the top sentence retrieved by our full model. For each image and reference sentence, phrases and top matching regions are shown in the same color. The matching score is given in brackets after each phrase (low scores are better). }
\label{fig:bidirectionalSuccess}
\end{figure*}

Table \ref{table:flickr30k}(c) shows results of this weighted distance with $\alpha = 0.7$. By itself, the performance of eq. (\ref{dis_ave}) is very poor, but when combined with $D_{CCA}(I,S)$, it gives a small but consistent improvement of 1\%-2\%. For completeness, the two lines of the Table \ref{table:flickr30k}(c) compare the performance of our full region-phrase model to just the basic VGG model. Despite big differences in R@1 for phrase localization (Table \ref{table:phraseloc}), the two models perform similarly for image-sentence retrieval. To understand why it is so difficult to get an improvement in image-sentence retrieval by incorporating increasingly accurate phrase localization models, it helps to examine retrieval results qualitatively. 

First, Figure~\ref{fig:bidirectionalSuccess} illustrates cases in which our region-phrase model does improve image-sentence retrieval performance.  In examples (a) and (b), the top retrieved sentences using the whole image-sentence model (left column) are incorrect but somewhat plausible.  However, the region-phrase model is unable to locate some the phrases from those sentences with any degree of confidence (e.g., {\em a checker} in (a), {\em people} in (b)).  However, the phrases of the correct sentences (right column) have much better region-phrase scores that compensate for the slightly worse whole image-sentence scores. The third example shows how our normalization term in eq.~(\ref{dis_ave}) helps longer sentences, which tend to have entities that are more difficult to localize.

\begin{figure*}
\centering
\begin{tabular}{rc|c}
& \large Correct sentence &  \large Top retrieved sentence\\
& \large  &  \large \\
\large\bfseries(a) & \multicolumn{1}{l|}{ Image-Sentence Score: 0.57}  & \multicolumn{1}{l}{Image-Sentence Score: 0.44} \\
 &\multicolumn{1}{l|}{Region-Phrase Score: 0.24}  & \multicolumn{1}{l}{Region-Phrase Score: 0.30} \\
&\includegraphics[width=0.45\textwidth,trim=0.2cm .2cm 0cm 0cm]{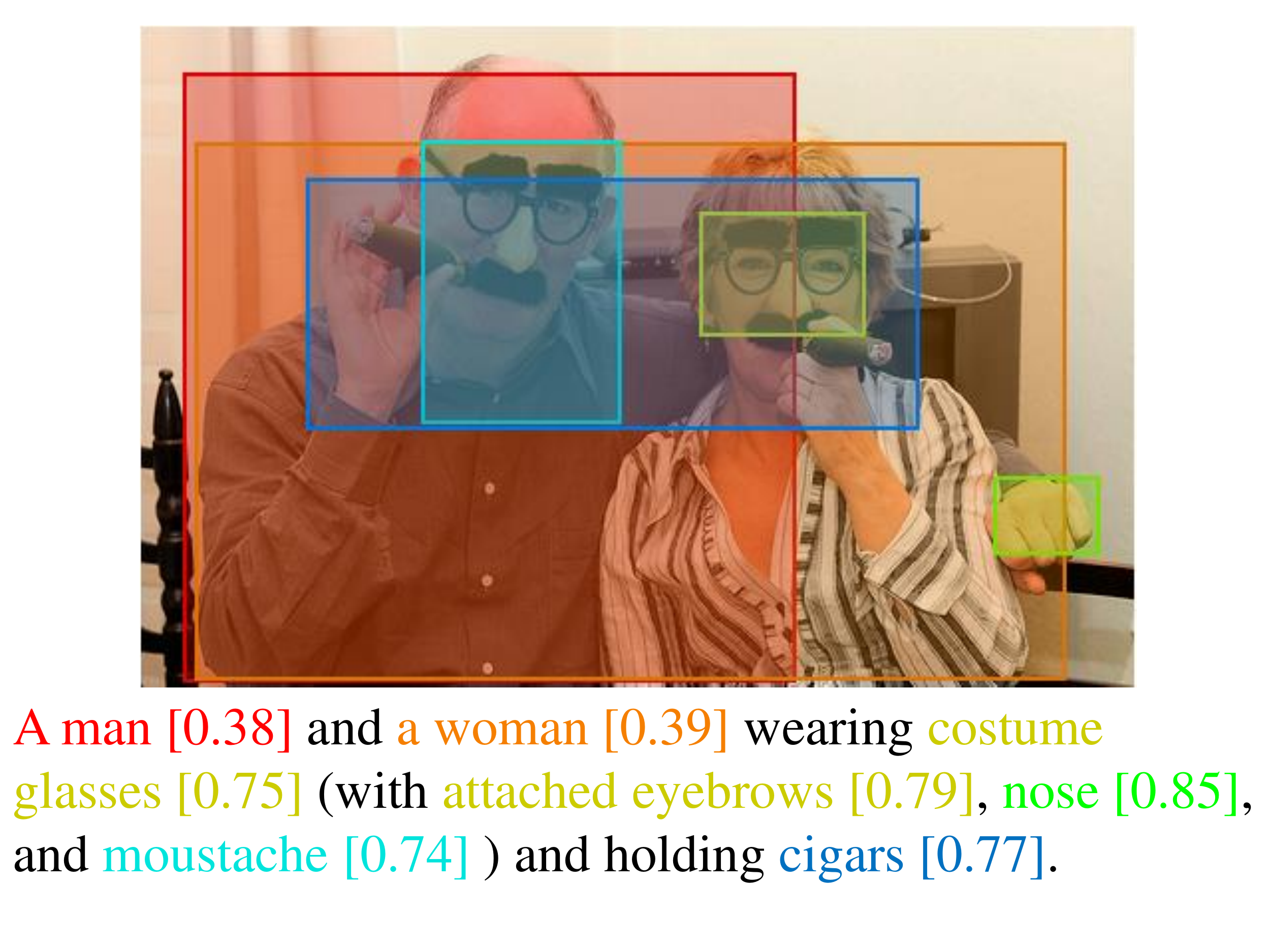} & \includegraphics[width=0.45\textwidth,trim=0.2cm .2cm 0cm 0cm]{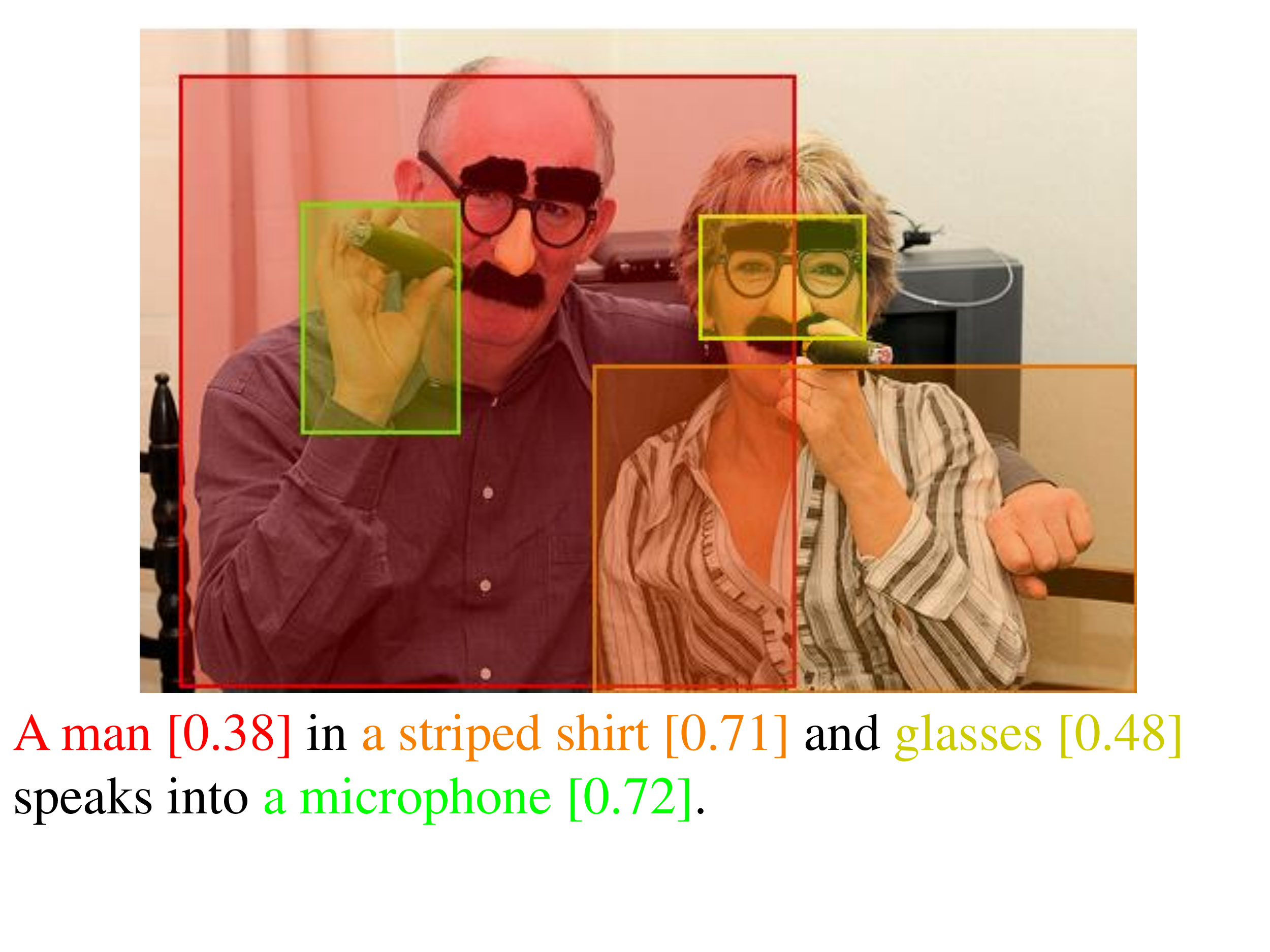} \\
\large\bfseries(b) & \multicolumn{1}{l|}{ Image-Sentence Score: 0.68}  & \multicolumn{1}{l}{Image-Sentence Score: 0.53} \\
 &\multicolumn{1}{l|}{Region-Phrase Score: 0.35}  & \multicolumn{1}{l}{Region-Phrase Score: 0.42} \\
&\includegraphics[width=0.45\textwidth,trim=0.2cm 1.5cm 0cm 0cm]{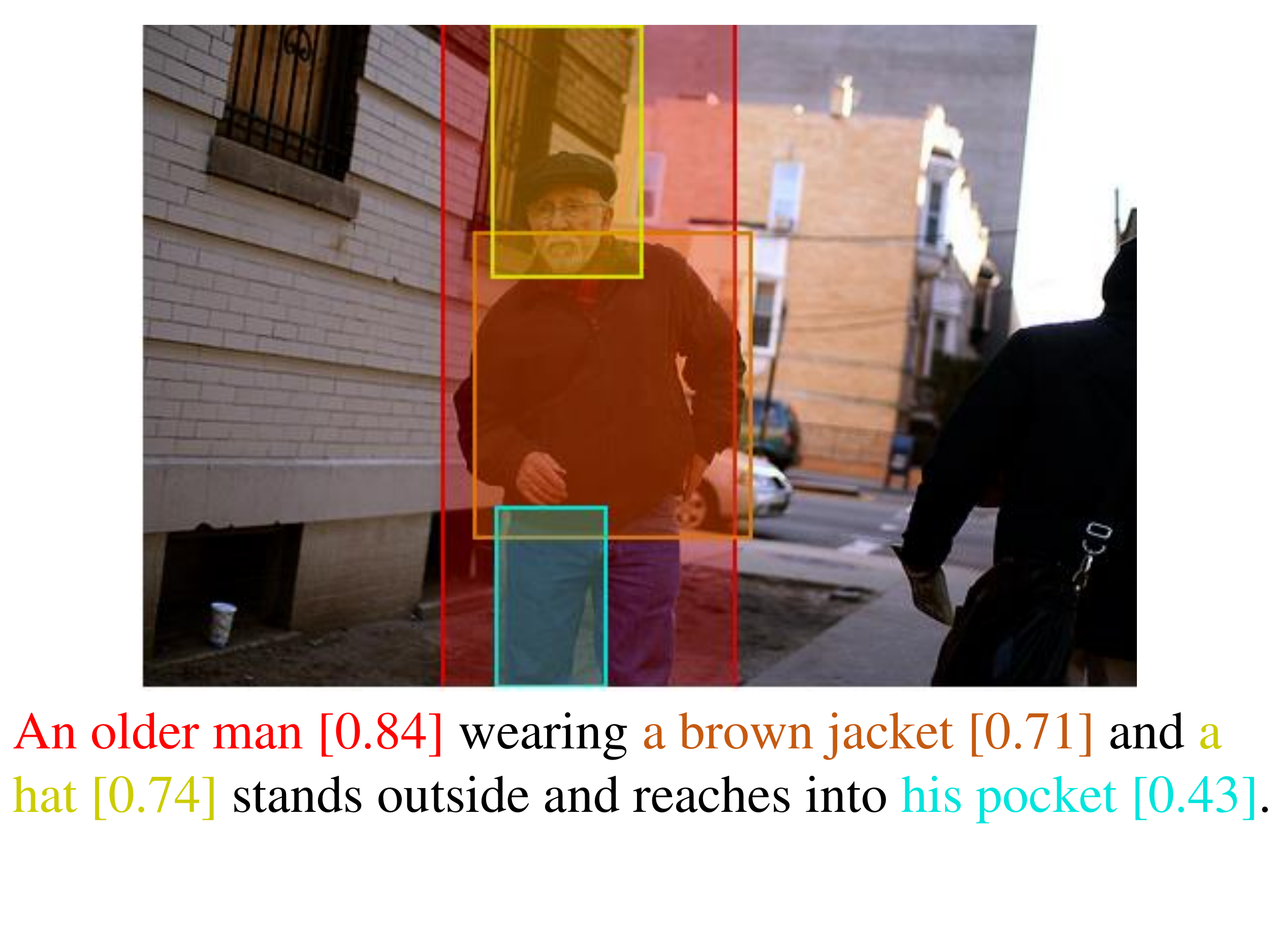} & \includegraphics[width=0.45\textwidth,trim=0.2cm 1.5cm 0cm 0cm]{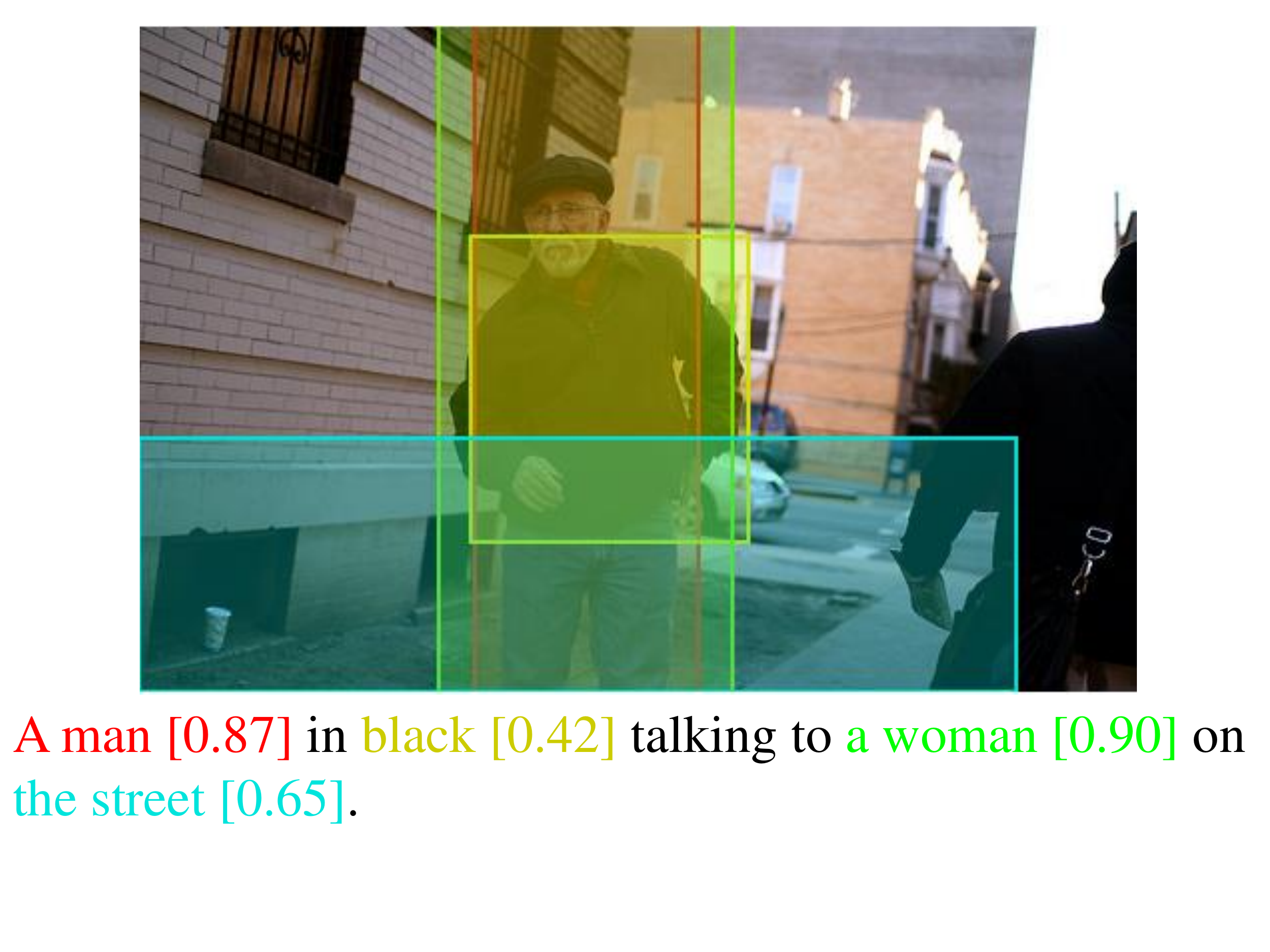} \\
\end{tabular}
	\caption{Example image-sentence retrieval results where region-phrase correspondences do not help to retrieve the correct sentence. For each test image, the left column shows a ground-truth sentence and the right column shows the top sentence retrieved by our method. For each image and reference sentence, phrases and top matching regions are shown in the same color. The matching score is given in brackets after each phrase (low scores are better). }
    \label{fig:bidirectionalOutput}
\end{figure*}

Despite the encouraging examples above, why is the overall quantitative improvement afforded by region-phrase correspondences so small? As we can see from the left column of Figure~\ref{fig:bidirectionalSuccess}, the global image-sentence CCA model usually succeeds in retrieving sentences that roughly fit the image. In order to provide an improvement, the region-phrase model must make fine distinctions, which is precisely where it tends to fail. Figure~\ref{fig:bidirectionalOutput} shows two examples of this phenomenon. For the first example image, the top sentence retrieved by our model includes a man, a striped shirt, and glasses, all with correct localizations in the image. There is also an incorrect, but plausible, localization of a microphone. However, our model is not discerning enough to figure out that the found instances of shirt and glasses do not belong to the man and that {\em a man and a woman wearing costume glasses} is a more accurate interpretation of the image than {\em a man with a striped shirt and glasses}. For the second example, the top retrieved sentence mentions a woman who is not there (and who our phrase localization model co-locates with the man). In order to make all of the above distinctions, we need not only a much more precise local appearance model, but a global contextual inference algorithm.
